Digital-Twin-Based Improvements to Diagnosis, Prognosis, Strategy Assessment, and Discrepancy Checking in a Nearly Autonomous Management and Control System

Linyu Lin, Paridhi Athe, Pascal Rouxelin, Maria Avramova, Nam Dinh

Department of Nuclear Engineering

North Carolina State University, Raleigh, NC

llin@ncsu.edu; pathe@ncsu.edu; pnrouxel@ncsu.edu; mnavramo@ncsu.edu; ntdinh@ncsu.edu

Abhinav Gupta

Department of Civil, Construction, Environmental Engineering

North Carolina State University, Raleigh, NC

agupta1@ncsu.edu

Robert Youngblood

Idaho National Laboratory, Idaho Falls, ID

robert.youngblood@inl.gov

Jeffrey Lane

Zachry Nuclear Engineering, Inc., Cary, NC

LaneJW@zachrygroup.com

**ABSTRACT**

The Nearly Autonomous Management and Control System (NAMAC) is a comprehensive control system that assists plant operations by furnishing control recommendations to operators in a broad class of situations. This study refines a NAMAC system for making reasonable recommendations during complex loss-of-flow scenarios with a validated Experimental Breeder Reactor II simulator, digital twins improved by machine-learning algorithms, a multi-attribute decision-making scheme, and a discrepancy checker for identifying unexpected recommendation effects. We assessed the performance of each NAMAC component, while we demonstrated and evaluated the capability of NAMAC in a class of loss-of-flow scenarios.

**ABBREVIATIONS**

| | |
|---|---|
| ANN | Artificial neural network |
| AOR | allowable operating region |
| BEP | best efficiency point |
| DT | digital twin |
| DT-D | diagnosis digital twin |
| DT-P | prognosis digital twin |
| EBR-II | Experimental Breeder Reactor II |
| FNN | feedforward neural network |
| GSIM | GOTHIC simulator |
| PFCL | peak fuel centerline temperature |
| POR | Preferred Operating Region |
| I&C | instrumentation and control |
| LOF | loss of flow |
| MSE | mean squared error |
| NAMAC | Nearly Autonomous Management and Control |
| PSP | primary sodium pump |
| RMSE | root mean square error |
| RNN | recurrent neural network |
| SCRAM | From the NRC Web site: "the sudden shutting down of a nuclear reactor usually by rapid insertion of control rods." (The etymology of the term is unclear.) |
| SMBO | sequential model-based optimization |
| SSF | safety-significant factor |

1. **INTRODUCTION**

Over the past few decades, there has been interest in the development of autonomous control systems for various real-world applications (e.g., aerial and ground transportation, industrial manufacturing, and more recently, nuclear reactors). Recent advancements in computer performance and artificial intelligence have played an important role in the advancement of autonomous control systems having a substantial degree of autonomy. The degree of autonomy for an autonomous control system is governed by its ability to accomplish a set of goals in the presence of a set of uncertainties with minimal or no external intervention [1] [2]. The uncertainties are due to external factors (e.g., environment) as well as internal factors (e.g., availability of data and computational models used in the development of autonomous systems). Therefore, the identification and systematic treatment of uncertainty become essential in the development of an autonomous control system.

In recent years, there has been interest in the development of autonomous control systems for nuclear reactors from researchers across the field. In 2003, R. Uhrig and L. Tsoukalas [3] stressed the importance of autonomous system in predicting the course of plant functionalities and acting in a manner that ensures their maintenance and well-being for generation IV reactor. Later, Na *et al.* [4] and Upadhyaya *et al.* [5] developed autonomous control systems for space reactors, respectively. K. Hu and J. Yuan [6] developed a multi-model predictive control method for nuclear steam generator water level by approximating the system behaviors with a polytopic uncertain linear parameter varying model. Darling *et al.* [7] developed a fault detection and management system for sodium fast reactors using dynamic probabilistic risk assessments and counterfactual reasoning. Cetiner *et al.* [8] developed a supervisory control system for the autonomous operation of advanced small modular reactors. Wang *et al.* [9] developed a state-space model predictive control method for core power control in pressurized water reactor based on a state-space representation of the reactor core and quadratic programming. More recently, with the advancement in machine learning and artificial intelligence, R. Coban [10] developed a power-level controller with neural network and particle warm optimization. Lee *et al.* [11] proposed a function-based hierarchical framework that uses long short-term memory for the development of autonomous control systems. Lee *et al.* [12] developed autonomous power-increase autonomous system with deep reinforcement learning and a rule-based system. Lin *et al.* [13] developed a Nearly Autonomous Management and Control (NAMAC) system for advanced nuclear reactors with digital twin (DT) technology. In that work, the demonstration of "proof of concept" was presented by implementing a baseline NAMAC, which derived reasonable recommendations for a specific loss of flow (LOF) transient, but the initial implementation was constrained by limited scalability. Although this is not the first time when machine learning algorithms or DTs are used in the autonomous control system, NAMAC is the first system that implements components of different functions with different techniques and combines them with a structured workflow to promote a rigorous, comprehensive, and realistic safety case. In addition, NAMAC stresses the importance of explainability of intelligent systems by deriving control recommendations from an interpretable and transparent knowledge base, including the operating procedures, issue spaces, reactor technical specifications, data warehouses, and a modular NAMAC architecture.

The NAMAC system [13] follows a three-layer architecture based on the knowledge base, developmental layer, and the operational layer. NAMAC adopts the modular design principle, where the goal is to enable transparency and to combine heterogeneous knowledge applications that interact with each other and with the complex environment. As a result, NAMAC contains DTs, both diagnosis DTs (DT-D) and prognosis DTs (DT-P), that uses the knowledge base for recovering full reactor states and for predicting the effects of mitigation strategies. Meanwhile, NAMAC has a strategy inventory for finding available mitigation strategies, strategy assessment for deciding the optimal strategy, and discrepancy checker for identifying unexpected action effects.

The development of the NAMAC system is an iterative process that is mainly driven by the uncertainty and scalability. For both separate components and integrated system, the uncertainty is measured in broad issue spaces that locate inside and outside the training domain. The scalability is measured by the consistency of NAMAC concept, including its architecture, component functions, development, and assessment process, from simplified to complex scenarios. This paper describes an implementation of the NAMAC system for situations involving higher complexity and uncertainties in the knowledge base. These complexities in the knowledge base are observed due to realistic pump controls, nonlinear responses, and other improvements in the data generation model to mimic the Experimental Breeder Reactor II (EBR-II) reactor behaviors more accurately. Compared to the baseline NAMAC in [13], the NAMAC system in this study features an improved DT modeling and strategy inventory, multi-attribute decision-making schemes, and a discrepancy checker. Meanwhile, separate component and integral system assessments are performed to demonstrate the flexibility and capability of NAMAC's modular framework in adapting to different sources of uncertainty.

This paper is organized as shown in Figure 1. Section 2 reviews the performance and findings from the NAMAC proof of concept, based on which section 3 discusses the requirements for improved DTs, components, and NAMAC system. Section 4 discusses the implementation of the knowledge base. As the foundation of the NAMAC development process, information from the knowledge base is used to develop the DT-D as described in Section 5.1, the DT-P in Section 5.2, the strategy inventory in Section 5.3, the strategy assessment in Section 5.4, and the discrepancy checker in Section 5.5. After the implementation of all required DTs and components, Section 6 discusses how these components are connected by an operational workflow. It also demonstrates the use of the NAMAC system on an EBR-II plant simulator.

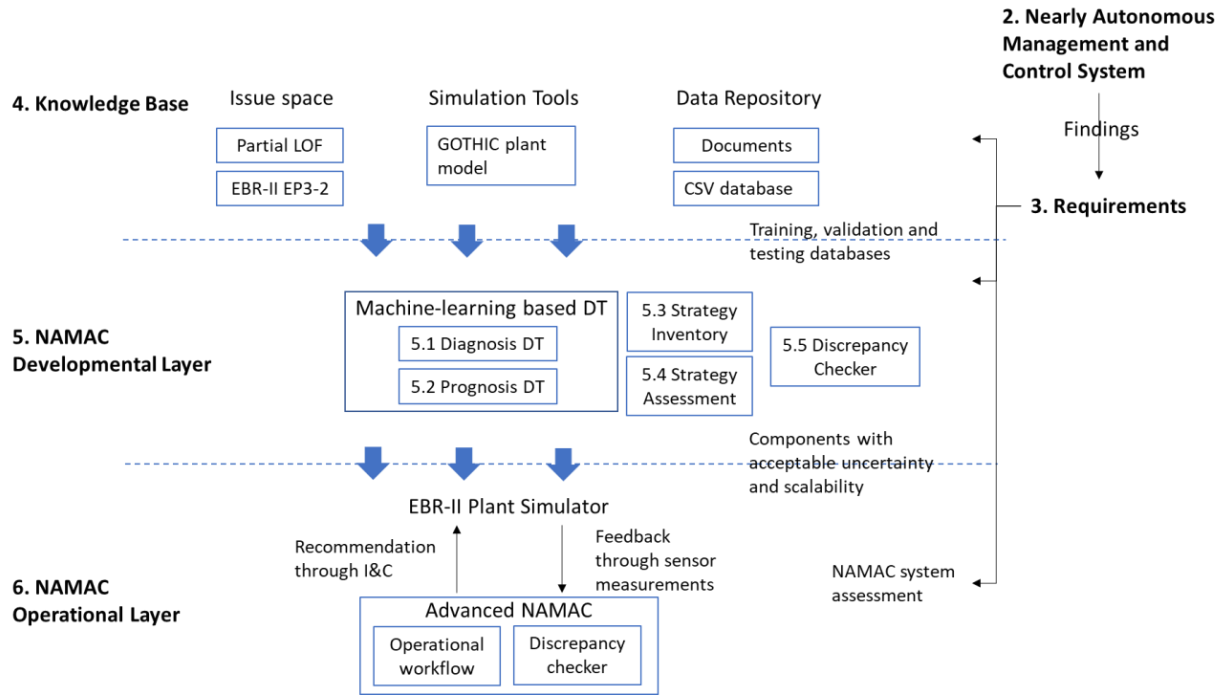

Figure 1: Scheme for the NAMAC development process. The number suggests corresponding sections that explain technical details for each key element.

## 2. NEARLY AUTONOMOUS MANAGEMENT AND CONTROL SYSTEM

The nearly autonomous management and control (NAMAC) system [13] is a comprehensive control system to assist plant operations by furnishing recommendations to operators. Such recommendations are derived by integrating knowledge from scenario-based model of plant, plant operating procedures, real-time measurements, etc. As an intelligent operational software, the goal of NAMAC is not to replace human operators, but to try to make use as much and as broader knowledge as possible, especially during emergent and difficult situations.

To better extract and store knowledge, NAMAC uses the DT technology, where we rely on the power of data-driven methods and machine learning algorithms in expressing highly nonlinear behaviors and heterogenous information during nominal and accident transients [14]. In NAMAC, the DTs are defined as a knowledge acquisition system from the knowledge base to support the intended uses. Specifically, each DT is defined by three attributes: function, interface, and modeling.

The DT function indicates the intended uses of a DT, and NAMAC contains two major DT functions: diagnosis and prognosis. The diagnosis is defined as a DT to "assimilate data from the operating plant to evaluate the complete states of a physical systems, including unobservable state variables, operating and fault conditions" [13]. Similar to fault detection and diagnosis methods reviewed in [15], DT-D in NAMAC also aims to identify operating modes, deviations, and causes of deviations by capturing correlations among state variables in different conditions based on safety-significant factors (SSFs). The prognosis is defined as a DT for "predicting the future transients of state variables or lifecycles of certain components based on the past histories and current information" [13]. Different from prognostic system for maintenance [16], DT-P in NAMAC estimates short-term system states due to control actions and strategies. DT-P provides insights equivalent to modeling and simulation but in a much faster path. Moreover, DT-P is tightly coupled with operational data and decision-making module, which potentially result in a different attractor than system simulations.

The DT modeling refers to methods and techniques for implementing DTs, including model-free methods or data-driven methods, model-based methods, and hybrid methods. The DT interface refers to the information transmission related to other components of control systems, human operators, and the physical system. In addition, NAMAC consists of strategy inventory for identifying available strategies based on the current states of the reactor, including the diagnosed plant state, safety limits, component control limits, etc.; the strategy assessment for ranking strategies based on the consequence factors, DT-P predictions, and a user-defined preference structure, which is created based on decision-makers' preference for safety, operability, and/or performance of the reactor.

Once implemented, these DTs and components are connected by a structured workflow (Figure 2). At the recommendation time, the DT-D first reads the sensor data from the simulator or real facility and restores the complete reactor states by predicting unobservable SSFs, like peak fuel and peak cladding temperature. These SSFs can be treated as surrogates of reactor's safety conditions, based on which potential failures and faults are identified [17]. In this study, DT-D only predicts the SSFs based on sensor readings since only one failure mode, i.e., partial LOF, is considered. Next, based on the DT-D predictions, the strategy inventory identifies all available mitigation strategies. Meanwhile, the DT-P performs front running predictions that evaluate the consequences of different strategies and initiating events. Later, these predictions are fed into the strategy assessment for deciding the optimal path that satisfies specific goals and preference. To stress the importance of AI trustworthiness, we have the discrepancy checker that continues monitoring and evaluating the differences between expectations and observations. Measured by certain metrics, if such differences are larger than a limit, the discrepancy checker decides that the DTs' outputs or NAMAC's recommendations can no longer be trusted either because it is operated outside the domain or there is too much uncertainty. Then, a safety-minded signal, for example, scram command, will be send to the operator.

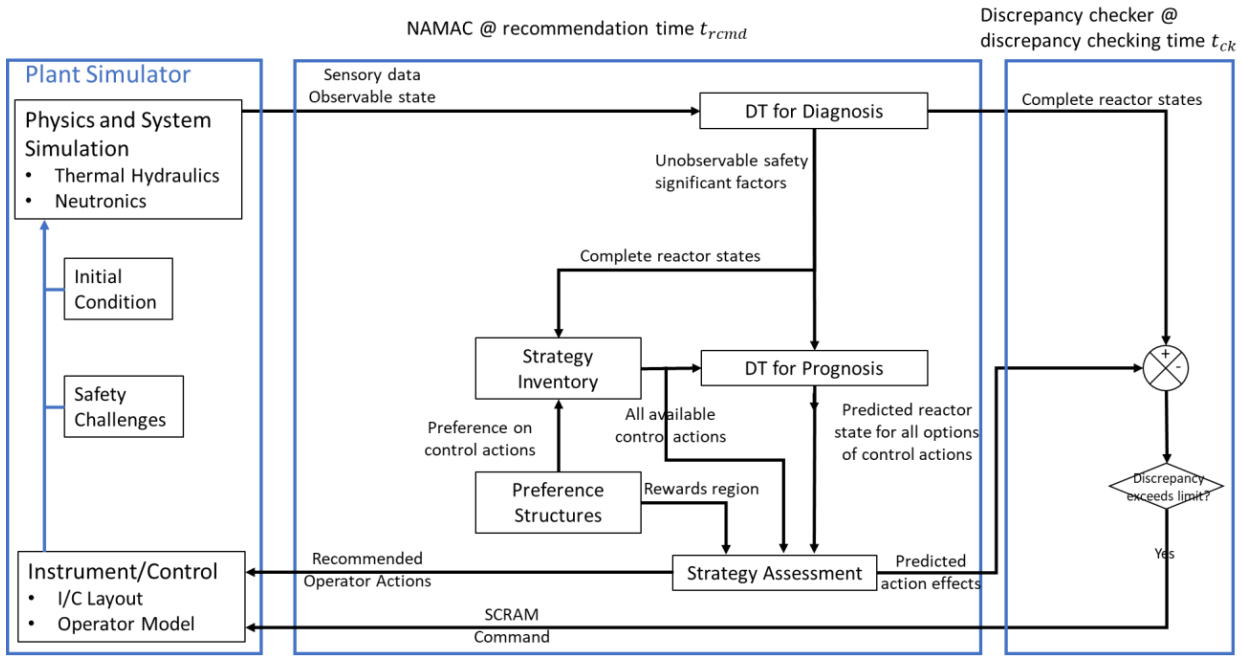

Figure 2: NAMAC operational workflow with discrepancy checker.

In [13], a baseline NAMAC was developed, as a proof of concept, to provide recommendations to the operator during a specific LOF scenario: starting from nominal operation, one of PSPs lost 50% rotational speeds in 50 sec. The database was obtained by sampling the scenario and control parameters, based on which DT-D, DT-P, strategy inventory, and strategy assessment where developed. We found both DT-D and DT-P had acceptable performance in the training domain. However, their prediction errors grown significantly, especially for DT-P, when DT-D and DT-P were operated outside the training domain. At the same time, we designed a preference structure that ranked strategy options based on safety margins. As a result, the strategy assessment determined the distance from achievable maximum peak fuel centerline (PFCL) temperatures to the safety limit, which was also characterized by PFCL temperature, and the best strategies were the one with largest distance. Moreover, we presented a case study for operating a simplified EBR-II simulator by NAMAC. We found that when the speed losses were maintained at 50% and when the NAMAC was activated for making recommendations at different recommendation time, the baseline NAMAC could make reasonable recommendations with small decision-making errors. However, when scenarios had speed losses higher or lower than 50%, the decision errors of baseline NAMAC grown largely.

To improve the performance of NAMAC and its components, this study aims to generate a knowledge base in more complex scenarios with a validated EBR-II simulator and multiple databases for the training and testing of NAMAC. Meanwhile, this study aims to investigate different modeling options for both DT-D and DT-P to ensure scalability and acceptable uncertainty in both training and testing domains. This study also aims to investigate model-based strategy inventory for predicting the trajectories of PSP torques and multi-attribute strategy assessment for making decisions based on multiple operational goals. Moreover, this study aims to improve the trustworthiness of NAMAC system by implementing a discrepancy checker, which detects unexpected reactor states and alerts operators for safety-minded actions. At last, this study aims to demonstrate the capability and performance of NAMAC in controlling an EBR-II simulator for a wide range of LOF scenarios with different magnitudes and speeds.

## 3. REQUIREMENTS

The development and assessment of a NAMAC system continuously reduce the uncertainty and improve the scalability for both separate components and the integral system. The uncertainty is measured by component or NAMAC errors in predicting the selected quantity of interest or in guiding the decision-making. Scalability is defined as the ability of the NAMAC concept, including its architecture, component functions, development, and assessment process, to be used in a range of target applications. References [18] [19] summarize three phases for NAMAC development: scoping, refinement, and maturation. From scoping to maturation, the uncertainty continues to be reduced by the best available methods and techniques. Meanwhile, scalability is repeatedly tested by evaluating the applicability and consistency of related methods and techniques. As a result, this study must refine NAMAC's components and system based on the success criteria on uncertainty. At the same time, this study aims to demonstrate the consistency of the NAMAC concept to increasingly complex scenarios.

Table 1 lists the requirements for uncertainty and scalability. The uncertainty requirements are designed based on three levels of success criteria for numerical demonstration [13], where the uncertainty of separate components and the integral system need to be evaluated. In this study, the root mean squared error (RMSE) metric $\varepsilon$ is used to quantify the errors, and the acceptance criteria for the uncertainty assessment are given by:

$$\varepsilon \leq \lambda \cdot QoI \qquad \text{Eq. 1}$$

where $QoI$ stands for the nominal values of quantities of interest. $\lambda$ refers to the uncertainty limit, and this study selects 5% based on previous studies [13] [14].

Table 1: List of requirements for the uncertainty and scalability of NAMAC separate components and integral systems.

|  | Uncertainty | Scalability |
|---|---|---|
| Component-wise requirements | | |
| **DT-D** | Errors in predicting safety-significant factors are acceptable | The component architecture, function, development and assessment process are consistent from simple to complex scenarios |
| **DT-P** | Errors in predicting the transient of selected variables are acceptable | |
| **Strategy inventory** | Errors in predicting the trajectories of control strategies are acceptable | |
| System-wise requirements | | |
| **Recommendation effects** | Errors in predicting the action effects for NAMAC recommendations are acceptable | The system operational workflow and reactor instrumentation & control interface, recommendations are consistent from simple to complex scenarios |
| **Decision rewards** | Errors in determining rewards for NAMAC recommendation are acceptable | |

In addition to DT-D, DT-P, and strategy inventory, this study also discusses the generation of the knowledge base and the implementation of two decision-making components: strategy assessment and the discrepancy checker. The scalability of the knowledge base is evaluated concerning the data generation process, while the decision-making components are evaluated concerning their transparency, flexibility, and scalability to different scenarios. However, at this stage of development, we did not consider the optimization of the decision-making scheme, including reducing uncertainty, refining constraints, eliciting and incorporating preference, etc.

## 4. KNOWLEDGE BASE

NAMAC relies on many different types of data and information coming from a wide variety of sources. Therefore, these "inputs" are the foundation of the NAMAC development and operation. Some examples of inputs include:

- Physical dimensions and geometry of the plant (or plant design).
- Systems, structures and components in the plant (or plant design).
- Constraints on SSCs (e.g., maximum pump speeds, maximum cool down rates, etc.) or other limits on the plant operational conditions (e.g., boiling or instability limits, power-to-flow maps, etc.).
- Control procedures and plant setpoints.
- Operational and transient data either from the operating plant or plant simulator including raw data from sensors, instruments, and controllers as well as record operator actions or other relevant data.
- Plant operating procedures and emergency operating procedures.
- Relevant experimental data.
- Available data from other plants of the same design (i.e., existing plants that are already operational).
- Definition of safety/operational/economic goals and "acceptable" risk.

Synthetic data from simulations is used at this stage of NAMAC development and this is consistent with what would be necessary for any first-of-a-kind (FOAK) reactor when it initially comes online. It is then expected that the synthetic data will be supplemented (or improved) over time based on actual operational history as measured by plant sensor data. However, it is expected that NAMAC will always need synthetic data throughout the lifetime of the plant. This is because:

a. The synthetic datasets produced by simulations tools are much richer datasets than are available from the plant itself. Plant sensors are limited both in the number and type of data that can be measured. Therefore, assimilating simulation-based data with measured plant data can provide access to information that was previously unobservable (e.g., peak cladding temperature).

b.  There will be design basis and beyond design basis transients that the plant does not endure and therefore operational data would not be available. Therefore, simulation based synthetic data sets will be necessary to fill in these knowledge gaps relative to the plant operating history.

It is also important to highlight here the different types of information that need to be considered by NAMAC, not just raw data from sensors or simulations. NAMAC also needs to consider procedures and other constraints on the operational domain. These represent valuable knowledge that needs to be considered by NAMAC to assist with quality of recommendations, computational efficiency and explainability of results.

The knowledge base stores information of three categories: issue space, simulation tool, and data repository. The issue space in this study comprises the flow anomalies ranging from partial to full LOF scenarios induced by various degrees of primary sodium pump (PSP) malfunctions. To mitigate the consequence of flow anomalies, this study refers to EBR-II Emergency Procedure 3-2, whose goal is to reestablish normal conditions given a primary flow anomaly. If the normal conditions cannot be reestablished, operator should scram the control rods, trip the primary pumps, and trip the primary-pump motor generator sets.

As a result, the NAMAC system will find actions that could restore the core flow by ramping up another PSP such that the reactor state is maintained in preferred operational regions. If no such action can be found or if the reactor behavior deviates substantially from the effects expected as a result of a NAMAC recommendation, the reactor will be Scrammed. Since there is a lack of operational data from nuclear reactors, especially in accident scenarios, this study uses a simulation tool named GOTHIC to model the thermal-hydraulic behaviors for the primary loop of EBR-II. GOTHIC is an industry-trusted, coarse grid multi-phase CFD tool that also includes the important attributes of traditional system level modeling tools such as component level models, control system capabilities and neutron point kinetics models [20]. GOTHIC applies a domain decomposition approach, allowing various levels of fidelity from 0-D to full 3-D to be applied in a single model, giving the user the ability to focus computational resources in the regions of interest while still capturing the integrated system response and important feedback effects. This makes GOTHIC a very computationally efficient tool for multi-scale applications and allows it to provide results that are unrealizable in a practical sense with either STH or CFD codes. The GOTHIC model of EBR-II used for the present work includes key physical features of the primary system with reactivity feedback effects, a representation of the intermediate loop that acts as a heat sink for the primary loop, and other plant control systems. GOTHIC has been benchmarked to EBR-II tests for loss of flow transients to confirm the evaluation model is suitable for this application [21].

To support the development of DTs and NAMAC components, the training, validation, and testing databases are generated by coupling GOTHIC with the Risk Analysis and Virtual ENvironment (RAVEN) framework [22]. By characterizing the accident scenario and mitigation strategy with mathematical forms, RAVEN draws samples from the issue space and runs GOTHIC simulations for a collection of transient behaviors of target state variables. These transients are stored in a comma-separated file (CSV) or HDF5 files, and databases with selected features will be used in the NAMAC development layer. In this study, the issue space is represented by mathematical formulations, where linear curves are used to describe the trajectories of pump malfunction and the corresponding mitigation strategies. As a result, the coverage of the issue space can be characterized by the malfunction magnitude, the beginning and ending of the malfunction, the magnitude of mitigations, and the beginning and ending of mitigations.. By changing control variables in RAVEN, we generated three databases with different ranges and numbers of samples for the selected issue-space characterizations. The mitigation strategies in database 1 (DB1) have different ramping speed for PSP torque trajectories, while DB2 and DB3 have a constant ramping speed. Meanwhile, compared to DB2, DB3 has different samples of magnitudes and starting time of mitigation strategies.

Table 2: Characteristics of three databases with different issue spaces.

|  | Database 1 (DB1) | Database 2 (DB2) | Database 3 (DB3) |
| --- | --- | --- | --- |
| **Malfunction magnitude (%)** | 26 samples ∈ [5–98] | 10 samples ∈ [0–98] | 6 samples ∈ [5–98] |
| **Mitigation magnitude (%)** | 26 samples ∈ [105–180] | 25 samples ∈ [100–150] | 16 samples ∈ [100–150] |
| **Mitigation starts (sec)** | 50 | 10 samples ∈ [50–100] | 13 samples ∈ [50–100] |
| **Mitigation ends (sec)** | 4 samples ∈ [80–130] | Starts + 50 | Starts + 50 |
| **Number of transients** | 4,804 | 5,000 | 2,496 |

In all three databases, each transient contains more than 400 variables extracted from GOTHIC. However, not all features are necessary for characterizing system responses. The selection of relevant state variables depends on the DT requirements and reactor instrumentation and control (I&C) system.

On one hand, from I&C concerns, pump torques are selected instead of pump speed since the use of speed would bypass the pump inertia model. For a flow anomaly induced by pump malfunctions and pump trips, setting the torque to zero and letting the pump coast down naturally is preferred and reasonable. To account for more severe conditions, large LOFs with rapid torque reductions and stuck pump are also considered in section 5 for NAMAC testing. As a result, for a pump-malfunction accident, a fractional decrease in nominal torque versus time is applied for one of the primary pumps, while the torque of another primary pump can be ramped up at corresponding rates. On the other hand, the DT requirements depend on the transients of state variables and characteristics of databases. Figure 3 shows the surface plot of fuel centerline temperature ($T_{PFCL}$) and core outlet flow rates against the torques of two PSPs. At normal operating conditions for 100% power and 100% flow, two PSPs are operated with torques equal to 636.57 N·m. Under a LOF scenario, PSP#1 can drop to 43.99 N·m due to malfunctions, while another PSP can ramp up to 963.58 N·m. In the designated issue space, the correlations between $T_{PFCL}$ or core outlet flow rate and the PSP I&C variables are highly nonlinear, especially when PSP#1 torques get smaller and the PSP#2 torques get larger. Figure 3 (a) shows that $T_{PFCL}$ increases drastically and distributes in a very different range when PSP#1 torques are less than 150 N·m. Such temperature peaks are induced by a large LOF, as shown in Figure 3 (b), with severe pump malfunctions. The nonlinearity in these low-torque regions is caused by the stuck pump, reverse flow through the weaker PSP, and the reactivity feedback modeled by point kinetics equations. As a result, to avoid safety consequences to core structures, the autonomous control system needs to monitor SSFs, including the peak fuel centerline (PFCL) temperature, peak cladding temperature, and core outlet flow rates. However, since it is difficult to measure fuel and cladding temperature directly by sensors, measurable variables that are closely correlated with SSFs (e.g., sodium temperature at the core inlet and outlet) are also needed.

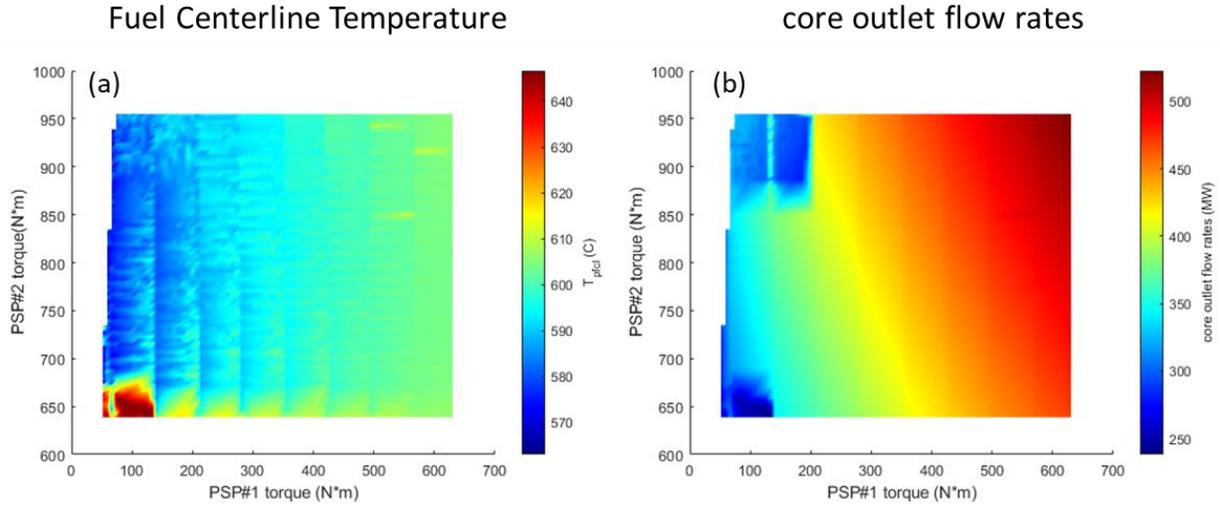

Figure 3: Surface plot of (a) fuel centerline temperature and (b) core outlet flow rates against the torques of two PSPs.

Meanwhile, considering the passive safety features of the EBR-II reactor, a temperature increase will automatically reduce the power generation rate without interfering with the power control systems [23]. In DB2, the average variation of core power rate is 12% of nominal power, while the average fuel temperature variation is 2%. As a result, for recoverable off-norm events, the mitigation strategies need not only to account for SSFs, like peak fuel and peak cladding temperature, but also to evaluate their economic impacts, including core power rates, intermediate heat exchanger power rate, and pump torques. The objective is to have NAMAC consider attributes beyond safety goals, including operational, economic, and system reliability, and maintenance goals. Details about the design of various operational regions from the perspective of safety, operation, and reliability are discussed in section 5.4. Table 3 lists relevant state variables selected from the knowledge base, together with their nominal values and relative ranges in DB2, that are used as the training databases for DT development. The specific NAMAC elements that each state variable is supporting are also listed, while their detailed uses are discussed in later sections.

Table 3: List of state variables extracted from the knowledge base and their ranges in DB2 to support the development and assessment of the NAMAC system.

| Relevant state variables | Nominal values | Ranges in DB2 | Supporting element |
|---|---|---|---|
| Core outlet flow rates | 469.8 kg/s | [50%–112%] | Diagnosis, prognosis |
| Upper plenum sodium temperature | 443.1°C | [99%–107%] | Diagnosis |

| High-pressure plenum sodium temperature | 344.4°C | [98%–100%] | Diagnosis |
|---|---|---|---|
| Low-pressure plenum sodium temperature | 344.4°C | [98%–100%] | Diagnosis |
| Pump torques | 636.6 N·m | [6%–150%] | Strategy inventory, prognosis, strategy assessment |
| Core power rates | 60.0 MW | [60%–110%] | Prognosis, strategy assessment |
| Intermediate heat exchanger power rates | 60.0 MW | [83%–103%] | Prognosis |
| PFCL | 605.8°C | [92%–107%] | Diagnosis, prognosis, strategy assessment |
| Peak cladding temperature | 487.9°C | [98%–115%] | Diagnosis, prognosis |

Overall, the complexity of LOF scenarios is represented by the nonlinearity of system behaviors, broad issue spaces, and the number and heterogeneity of state variables that are investigated by the NAMAC system. Meanwhile, the passive features of EBR-II reactors require NAMAC to account for more attributes than SSFs, including power variation, torque viability, etc., such that an optimal strategy can be found in a transparent and flexible manner. Moreover, we used the same data generation engine as the proof-of-concept demonstration [13], while the number of transients has increased from 2,000 to 12,000. As a result, the scalability of the knowledge base is ensured by a consistent development and assessment process for the knowledge base when the scenarios are transferred from simplified proof of concept to complex LOF accidents.

## 5. DEVELOPMENT OF A NAMAC SYSTEM FOR COMPLEX LOF SCENARIOS

Guided by the NAMAC development process (Figure 1), this section discusses the implementation of DT-Ds and DT-Ps, strategy inventory, strategy assessment, and the discrepancy checker.

### 5.1. Diagnosis Digital Twin

In the NAMAC system, the DT-D aims to assimilate data from the operating plant and knowledge from the knowledge base for recovering the target states of a reactor system. The workflow of DT-D is discussed in [8], and the same scheme is used in this study. Eq. 2 shows a general form of DT-D: $f_{DT-D}$ is the data-driven model of DT-D; $X_D(t) = [x_1(t), x_2(t), ..., x_N(t)]$ represents input variables of the DT-D model at time $t$, and $N$ is the number of input features; $P_D = [p_1, p_2, ..., p_M]$ represents training and design hyperparameters of DTs, including number of neurons, number of layers, batch sizes, learning rates, sequence length, regularization weights, validation patience, epoch number, and target training errors. $M$ is the number of hyperparameters; $KB_D$ represents the knowledge base used to train the data-driven models, including training databases, user knowledge in feature selections and preprocessing; SSF includes the peak fuel and peak cladding temperatures, and $\widehat{SSF}$ is predicted dynamically at every time point $t$; $\varepsilon_D$ represents the mean squared error (MSE) of the DT-D model in predicting the corresponding SSF. This study calculates MSE $\varepsilon_D$ using Eq. 3, where N is the number of predictions; and $SSF$ represents real values from the simulator.

$$\widehat{SSF}(t) = f_{DT-D}(X_D(t), P_D, KB_D) \qquad \text{Eq. 2}$$

$$\varepsilon_{MSE} = \frac{1}{N} \sum_{i=1}^{N} (SSF - \widehat{SSF}) \qquad \text{Eq. 3}$$

In the previous NAMAC system [13], a correlation between selected sensor measurements and the target SSFs was constructed by feedforward neural networks (FNNs). Sensitivity analysis on a list of sources of uncertainty was performed, and it was found that the database coverage, input errors, and model hyperparameters $P_D$ are found to have significant impacts on the diagnosis performance. In order to improve the accuracy of DT-D against these high-impact sources of uncertainty, this study also investigates the recurrent neural network (RNN), considering its capability in handling time sequences and in capturing nonlinear behaviors in complex thermal-hydraulic transients.

The FNN and RNN are both popular types of artificial neural networks (ANNs) that have been widely used in image recognitions, regression, classifications, etc. The neural network is composed of three elements: neurons, connections and weights, hyperparameters:

(1) The neurons are basic elements that receive inputs, combine inputs with their internal state, and produce outputs.
(2) The connections and weights are ways to connect the output of one neuron as an input to another neuron.
(3) The hyperparameters are constant parameters that are set before the learning process.

FNN was the first and simplest type of ANN, where the information flows in only one direction: from the input nodes through the hidden nodes and to the output nodes. For the class of networks with more than one layers, the universal approximation theorem states that "every continuous function that maps intervals of real numbers to some output interval of real numbers can be approximated arbitrarily closely by a multi-layer network with just one hidden layer" [24]. For RNNs, the neurons are connected to each other by recursive relationships in a hidden layer. As a result, RNN not only takes input data but also information at previous sequence steps into account such that the sequence data is considered for generating outputs. To account for sequence information, RNN needs to account for the whole sequence of input data, and the length of input sequence is a hyperparameter that is designated by users. Typical RNN neurons include gated recurrent units, long short-term memory units, etc. Figure 4 shows a typical scheme of RNN, where a feedback loop is included for allowing information to go through the same network. When the feedback loop is unfolded into a graph without any cycles, output $Y_t$, internal state $S_{t-1}$ from the previous sequence step, and the input data at the current sequence step $x_t$ are passed to the network as inputs for processing the next step [25]. The process enables the hidden layer to memorize all information that is learned from both the input data and previous states [26].

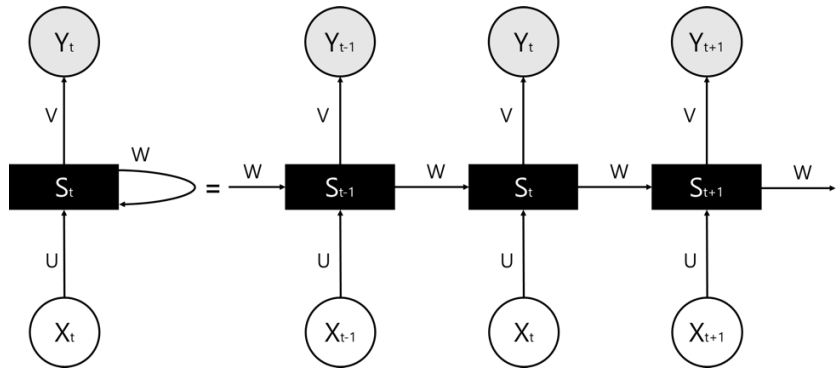

Figure 4: Illustration RNN structure.

where $x_t$ is the input at sequence step t, $S_t$ is the hidden state at sequence step t, $Y_t$ is the output at sequence step t. Eq. 4 and Eq. 5 describe how parameters (U, V, W), input $x_t$, and the hidden state at previous sequence step $S_{t-1}$ are used in the activation function ($f$):

$$S_t = f(Ux_t + WS_{t-1})$$ Eq. 4

$$Y_t = VS_t$$ Eq. 5

RNN has been used to model dynamic behaviors in complex systems, including robot control, time-series prediction, time series anomaly detection, etc. Compared to FNN, the RNN is designed to handle sequence and thus more suitable for forecast and prognosis functions.

A typical method to train an ANN is through gradient descent method, which minimizes the error term by changing weight inside each neuron in proportion to the derivative of the error with respect to that weight. In this study, the error term is calculated by MSE. The weight update process is conducted recursively until the stopping criteria are met, for example, the maximum number of iterations, i.e., epoch number, the values of error terms, i.e., the target training errors. In each iteration, the number of training data used is specified by batch sizes, and a batch size greater than one but less than the total data sizes is known as the mini-batch mode. Meanwhile, the number of weights in each computing process that will be updated is known as the learning rate. Learning rate is a configurable hyperparameters with a small positive value often in the range between 0 and 1. In this study, the learning rate is reduced when the validation errors stop decreasing for a designated number of iterations. The goal is to have the learning algorithms get closer to the minimal point with smaller steps and to result in higher accuracy.

Although ANN with multiple layers can approximate a wide variety of functions with arbitrary precision measured in $L^1$ or $L^\infty$ norm, a well performed ANN model on the training data can fail when tested on unseen data points. Such a problem is known as overfitting, and the ability of ANNs in predicting unseen data point is defined as the generalization capability. To prevent the problem of overfitting, regularization is one of the common techniques to reduce the complexity of the model in a manner that forces to improve ANN's generalization capability. In this study, L2 regularization is used by adding weights to the error term of ANNs. Regularization weight is a user-defined parameter that is directly proportional to the amount of regularization applied. The weight ranges from 0 to 1, where 1 means maximum regularization, while 0 means no regularization.

With designated hyperparameters, two DT-Ds (based on FNN and RNN, respectively) are trained using the same training databases and assessed by three testing databases (Table 4). Their assessment results are compared correspondingly to demonstrate the capability of RNN modeling option. The sodium temperatures at three locations (high-pressure plenum $T_{HPP}$, low-pressure plenum $T_{LPP}$, and upper plenum $T_{UP}$) are used as inputs $X_D$ to both DT-D models. Both the FNN and RNN have two layers, and each layer has 30 neurons. The training database contains 1,024 different transients: one primary pump is operating with 50% of nominal conditions, and such a malfunction is mitigated by ramping up another pump by 32 different magnitudes starting from 32 different times. During the training process, 80% of episodes are randomly selected from the training databases for updating weights and biases of neural network, while each episode contains 250sec transients for all selected state variables. The remaining data are divided in half and used as validation and testing data, respectively. The validation data is used to reduce the learning rate if validation errors stop decreasing for a designated number of iterations. The testing data is to stop the learning algorithm from minimizing the error terms if the testing errors stop decreasing for a designated number of iterations. In addition, three testing databases with different coverages on the issue space are generated to evaluate DT-D's uncertainty and generalization capability. In this study, the generalization capability is measured by RMSE in predicting different data points than the training data. Tests 1 and 3 correspond to testing scenarios that are outside the DT-D training domain with very different magnitudes of malfunction. Meanwhile, Test 2 has transients similar to those in training; the magnitude of malfunction and the range of mitigation are the same, but the number of samples for the magnitudes of mitigation is different. As a result, transients in Test 2 can be considered as interpolated within the range of training transients. By observing the MSE in predicting the target quantity, it is found in Table 4 that for both FNN and RNN approach, DT-D errors in Test 2 are similar to those in training. Although both DT-D's errors grow on Test 1, the RNN error is much smaller. It is also found that the FNN error grows significantly for Test 3, while the RNN's error remains at a similar order of magnitude to the training errors. We expect RNN-based DT-D to have smaller errors because of the time-sequence nature of training and test databases. By capturing the correlations in data sequence with the feedback loop, RNN makes more accurate predictions in training domain and all three tests. However, FNN neglects the time dependency and only captures the correlations between input and output variables. As a result, FNN only has good accuracy in the training domain or test 2 with similar data characteristics.

Table 4: Comparisons of DT-D errors with different coverage conditions.

|  | **Magnitude of malfunction** | **Magnitude of mitigation** | **DT-D errors (MSE)** | |
|---|---|---|---|---|
|  |  |  | **FNN** | **RNN** |
| **Training** | 50% | 32 samples from 100% to 150% | 2.57 | 1.94 |
| **Test 1** | 25% | 16 samples from 100% to 150% | 18.96 | 4.93 |
| **Test 2** | 50% |  | 2.62 | 1.94 |

| Test 3 | 75% | | 23.33 | 1.41 |

As another high-impact source of uncertainty, random noises are added to the transient of three inputs of DT-D based on Eq. 6. At each time $t$, the noise is independently and identically drawn from a normal distribution with zero mean and a standard deviation. The standard deviation is assumed to be a constant fraction of the real input $X_D$. In this study, the constant fraction $C$ equals 0.1%, and it is applied to all four databases in Table 4.

$$\hat{X}_D(t) = X_D(t) + \mathbb{N}(0, CX_D) \qquad \text{Eq. 6}$$

Table 5 compares the prediction errors of FNN-based and RNN-based DT-D when random noises are added to each input, respectively. Data shows that, when 0.1% of noises are added to $T_{LPP}$ and $T_{HPP}$, the RNN has smaller errors than FNN. However, RNN errors grow significantly when noises are added to $T_{UP}$ because RNN is more sensitive to the gradient of data sequence than FNN, and the sodium temperature at core outlet has higher gradients than inlet temperatures.

Table 5: FNN and RNN DT-D prediction errors when there are random noises in individual inputs.

| Inputs with random noise | $T_{LPP}$ | | $T_{HPP}$ | | $T_{UP}$ | |
|---|---|---|---|---|---|---|
| DT-D prediction errors | MSE by FNN | MSE by RNN | MSE by FNN | MSE by RNN | MSE by FNN | MSE by RNN |
| Training | 59.21 | 33.32 | 44.55 | 5.88 | 3.10 | 192.76 |
| Test 1 | 70.37 | 25.64 | 48.44 | 7.61 | 19.42 | 136.01 |
| Test 2 | 58.21 | 32.18 | 45.10 | 5.76 | 3.18 | 189.86 |
| Test 3 | 23.85 | 14.54 | 72.59 | 3.11 | 23.85 | 72.72 |

As a result, compared to FNN DT-D, the RNN model tends to be more scalable concerning the magnitude of malfunctions. Moreover, the RNN model is more robust against random noises added to the sodium temperature measurements at lower-pressure and high-pressure plenum, while the FNN model tends to be sensitive to random noises on all three inputs. Moreover, from extensive case studies [17], we observe that the RNN-based DT-D is able to accurately predict SSFs with four physical variables when there are designated sensor drifts and random noises. Considering the scalability and robustness requirements by the NAMAC development and operation, we select RNN-based DT-D for the NAMAC system.

To evaluate the performance of RNN-based DT-D, a list of sources of uncertainty is specified, while samples are drawn from the designated ranges. Table 6 lists sources of uncertainty, parameter ranges, and findings for the RNN-based DT-D. Each model fit parameter is sampled independently from the designated parameter range, and the strength of the correlation is quantified by the Pearson correlation coefficient (PCC) between the DT-D errors and the sampled parameters. Data shows a strong correlation when the absolute value of PCC is greater than 0.5. Our findings show that errors of RNN-based DT-D are strongly correlated with batch size, regularization weights, and database coverage in the designated ranges.

Table 6: A list of sources of uncertainty, parameter ranges, selected values, and findings for RNN-based DT-D.

| Category | Sources of uncertainty | Parameter range | Selected Value | Findings |
| --- | --- | --- | --- | --- |
| **Model fit** | | | | |
| | Sequence length | [5–20] | 5 | DT-D errors are slightly affected by sequence length in the designated range. |
| | Batch size | [100–600] | 100 | DT-D errors are strongly affected by batch size in the designated range, where smaller batch sizes result in smaller errors. However, training with small batch sizes could take longer time to reach the same training targets. |
| | Number of layers | [2–4] | 2 | DT-D errors are slightly affected by the number of neurons in the designated range. |
| | Regularization weight | $[0-10^4]$ | 0 | DT-D errors are strongly affected by regularization weight in the designated range, where smaller weights indicate smaller errors. However, small regularizations could greatly limit the generalization capability. |
| **Scope compliance** | | | | |
| | Data Coverage | DB1 DB2 DB3 | / | DT-D errors are strongly affected by data coverage, where similar databases result in smaller errors. |

To better identify the correlation between data coverage and DT-D errors, the similarities between the training and three testing databases are quantified by two mutual information metrics: symmetric Kullback-Leibler divergence and Hellinger distance. First of all, the transient data from the training and testing databases are fitted by kernel functions (Eq. 7).

$$\hat{f}_h(x) = \frac{1}{n}\sum_{i=1}^{n} K_h(x - x_i) = \frac{1}{nh}\sum_{i=1}^{n} K\left(\frac{x - x_i}{h}\right) \qquad \text{Eq. 7}$$

where $x$ represent the multidimensional random vectors with density function $P_i$; $x_i$ represents a random sample drawn from the database; $n$ represents the total number of data points drawn from the database; $K(x)$ is the kernel function, and this study uses the normal distribution; $h$ is the bandwidth, and the optimal bandwidth is determined by minimizing the mean integrated squared error [27]. Figure 5 shows the fitted distributions for the training databases, testing database DB1, DB2, and DB3 as in Table 2.

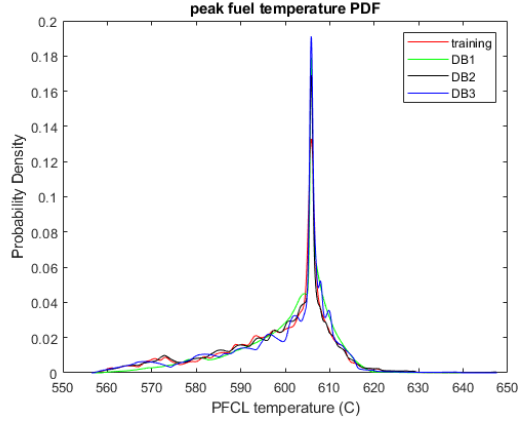

Figure 5: Probability distributions for the training databases, testing database DB1, DB2, and DB3. The distributions are fitted by kernel functions.

Next, the mutual information is calculated with the symmetric Kullback-Leibler (K-L) divergence (Eq. 8 & Eq. 9) and Hellinger distance (Eq. 10) separately.

$$D_{KL}(P||Q) = \sum_{x \in X} P(x) \log\left(\frac{P(x)}{Q(x)}\right) = \int_{-\infty}^{\infty} p(x) \log\left(\frac{p(x)}{q(x)}\right) dx \qquad \text{Eq. 8}$$

$$D_{S-KL}(P, Q) = D_{KL}(P||Q) + D_{KL}(Q||P) \qquad \text{Eq. 9}$$

$$H^2(P, Q) = \frac{1}{2}\sum_{x \in X}\left(\sqrt{P(x)} - \sqrt{Q(x)}\right)^2 = \frac{1}{2}\int\left(\sqrt{dP} - \sqrt{dQ}\right)^2 \qquad \text{Eq. 10}$$

To determine the strength of correlations between the mutual information results from the symmetric K-L divergence and Hellinger distance and the DT-D errors, the PCC is used to determine the degree of linear correlations between the mutual information results and the DT-D errors (Eq. 12), which is calculated by RMSE as in Eq. 11.

$$\varepsilon_{RMSE} = \sqrt{\frac{1}{N}\sum_{i=1}^{N}(\hat{y}_i - y_i)^2} \qquad \text{Eq. 11}$$

$$\rho_{KL,\varepsilon_{RMSE}} = \frac{cov(M, RMSE)}{\sigma_{KL}\sigma_{\varepsilon_{RMSE}}} \qquad \text{Eq. 12}$$

where $\hat{y}_i$ is predictions for data point $i$ by DT-D; $y_i$ is real values from the database for the same data point $i$; $cov(KL, RMSE)$ represents the covariance of mutual information results, the testing RMSE $\varepsilon_{RMSE}$, $\sigma_{KL}$ represents the standard deviation of symmetric K-L divergence, which can be replaced by $\sigma_H$ if the metric is Hellinger distance, and $\sigma_{\varepsilon_{RMSE}}$ is the standard deviation of DT-D prediction errors quantified by RMSE $\varepsilon_{RMSE}$. Table 7 shows the mutual information results of three databases relative to the training database. For the training database itself, mutual information equals 0, which indicates that the distribution is the same and that both mutual information metrics have been correctly implemented. The training error is 0.02°C. Since the training data is randomly sampled from DB2, Test 2 is the most similar to the training data and thus has the smallest symmetric K-L divergence and Hellinger distance. The DT-D errors are also the smallest, as the RNN is predicting data points that are similar to the training domain. DB1 has the largest K-L divergence and Hellinger distance as the durations of mitigation strategies are sampled, in contrast to the constant duration in DB2 and DB3. DB3 has the same issue space as the training data. Since DB3 is generated by different samples from DB2, the divergence and distance results are smaller than DB1 vs. training and larger than DB2 vs. training. Meanwhile, the DT-D errors are comparative in predicting DB1 and DB3 because of a better generalization capability by the RNN modeling option. Moreover, the PCC is 0.54 for symmetric K-L divergence and 0.69 for Hellinger distance, which both suggest strong correlations.

Table 7: Mutual information of three testing databases to the training database. The PCCs are calculated between the DT-D errors and the mutual information results.

|  | Symmetric K-L divergence | Hellinger distance | DT-D errors |
|---|---|---|---|
| **Training database** | 0 | 0 | 0.02°C |
| **DB1** | $6.6 \times 10^{-1}$ | $4.1 \times 10^{-1}$ | 0.34°C |
| **DB2** | $1.9 \times 10^{-2}$ | $9.2 \times 10^{-2}$ | 0.14°C |

| DB3 | $6.1 \times 10^{-2}$ | $4.8 \times 10^{-2}$ | 0.35°C |
| PCC | 0.54 | 0.69 | |

Overall, the improvement of RNN-based DT-D is reflected by the acceptable uncertainty when it is used both inside and outside the training domain. The RNN-based DT-D is found to be more robust against the designated random sensor noises than FNN. The scalability of DT-D development and assessment process are demonstrated by using the same formulation and workflow to implement both RNN-based and baseline FNN-based DT-D, which results in comparable performance in the training domain. Through sensitivity analysis, this study identifies strong correlations between the DT-D errors and the selected hyperparameters. Moreover, we observed a strong correlation between the DT-D errors and the data coverage conditions, quantified by mutual information, while we learned similar lessons from the NAMAC proof of concept.

### 5.2. Prognosis Digital Twin

To find the optimal actions, an accurate prediction of action effects is required. A DT-P is built to predict, for each mitigation strategy option, the future transients of selected state variables based on the operating histories and the diagnosis. The workflow of DT-P is discussed in [8], and the same scheme is used in this study. Eq. 13 shows the general form of DT-P: $f_{DT-P}$ is the data-driven model for DT-P; $P_P$ is the hyperparameter of data-driven model $f_{DT-P}$; $KB_P$ is the knowledge base for DT-P training; $A(t) = \begin{bmatrix} a_{11}(t) & a_{12}(t) & \dots & a_{1M}(t) \\ a_{21}(t) & a_{22}(t) & \dots & a_{2M}(t) \\ \vdots & \vdots & \ddots & \vdots \\ a_{P1}(t) & a_{P2}(t) & \dots & a_{PM}(t) \end{bmatrix}$ contains in total $P$ options of mitigation strategies, while each option is composed by $M$ actions; and $X_P(t) = \begin{bmatrix} x_{11}(t) & x_{12}(t) & \dots & x_{1N}(t) \\ x_{21}(t) & x_{22}(t) & \dots & x_{2N}(t) \\ \vdots & \vdots & \ddots & \vdots \\ x_{P1}(t) & x_{P1}(t) & \dots & x_{PN}(t) \end{bmatrix}$ is the selected set of state variables at time $t$ for all options of mitigation strategies, $N$ is the number of variables, where the future states $X_P(t + \Delta t)$ are predicted base on the history and current information $[X_P(t_1), X_P(t_2), \dots, X_P(t)]$.

$$X_P(t + \Delta t) = f_{DT-P}(A(t), [X_P(t_1), \dots, X_P(t)], P_P, KB_P) \qquad \text{Eq. 13}$$

There are different modeling options to implement DT-P. In the previous implementation [13], the FNN was used to directly predict the consequence factor based on the mitigation strategy options and history information. In the present study, RNN is used to perform multistep predictions such that the entire transients of selected state variables $X_P$ can be derived. Figure 6 shows the scheme of RNN-based DT-P for multistep predictions. Before the prediction/recommendation time $t_r$, the hidden states $s_t$ in the RNNs are updated based on the sensor measurement of $X_P(t_0 \sim t_r)$ and control histories $A(t_0 \sim t_r)$. After time $t_r$, the hidden states will be updated based on the assembled inputs from the DT-P predictions at the previous step and the predicted trajectory of mitigation strategy from the strategy inventory. The prediction will be conducted recursively until the designated end of the transient $t + T$.

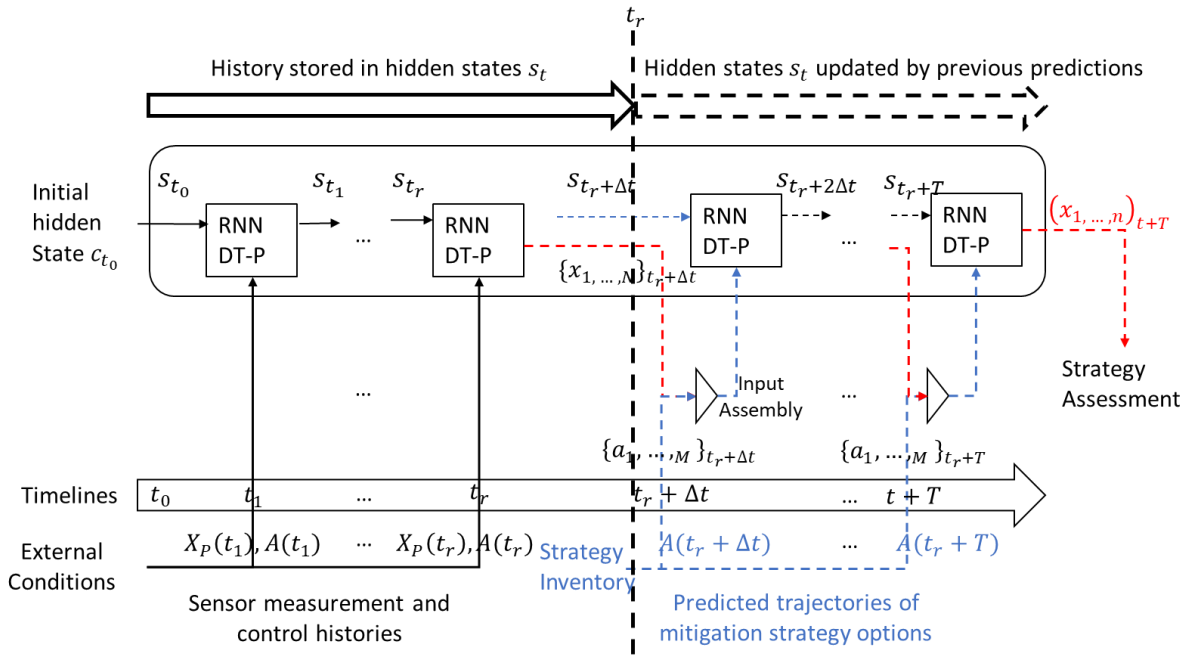

Figure 6: Scheme of RNN-based DT-P for multistep predictions.

Figure 7 (b) compares the $T_{PFCL}$ predictions by RNN DT-P and FNN DT-P against real values from the database in response to one of the mitigation strategies. The mitigation strategies are represented by torque transients for both PSPs, which are plotted in Figure 7 (a). The FNN only predicts the highest fuel temperature as the consequence factor for each mitigation strategy, while the RNN is able to predict the entire transients for selected variables. As a result, RNN could better assess action effects and support decision-making with more comprehensive information.

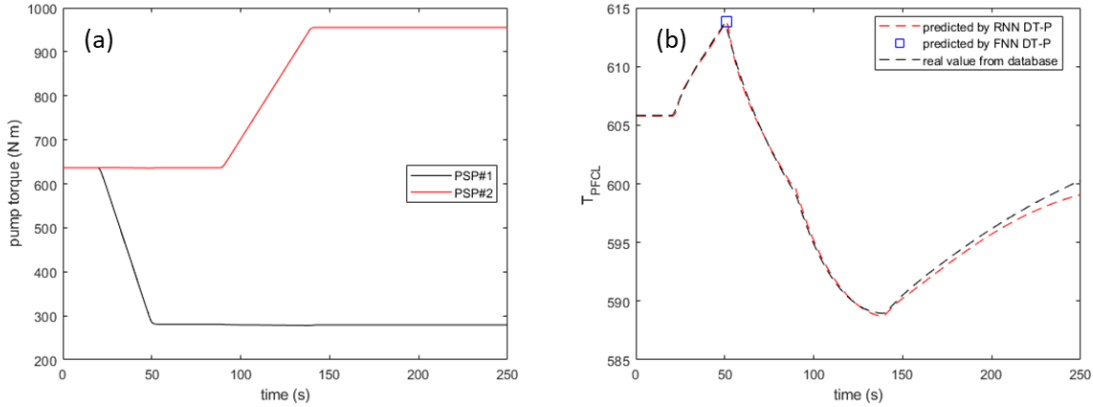

Figure 7: (a) The transient of torques for PSP#1 and PSP#2 as predicted by the strategy inventory; (b) comparison of fuel centerline temperature as predicted by RNN-based DT-P, FNN-based DT-P, and real values from databases in response to the torque transients in (a).

Different from the FNN model, RNN requires a certain length of history $X_P(t_h \sim t_r)$, defined as the starting time $t_h$ for updating the hidden states, such that the future transient of state variables can be accurately predicted. Figure 8 compares the DT-P predictions for (a) one transient of $T_{PFCL}$ and (b) all 5,000 transients of $T_{PFCL}$ as described in Section 4 with 0 s, 5 s, and 20 s, respectively. The errors grow significantly when no history information is available. For all 5,000 transients, the MSE by Eq. 3 is 2,219 when there is a 0 s history, and it is reduced to 9.9 when there is a 5 s or 20 s history.

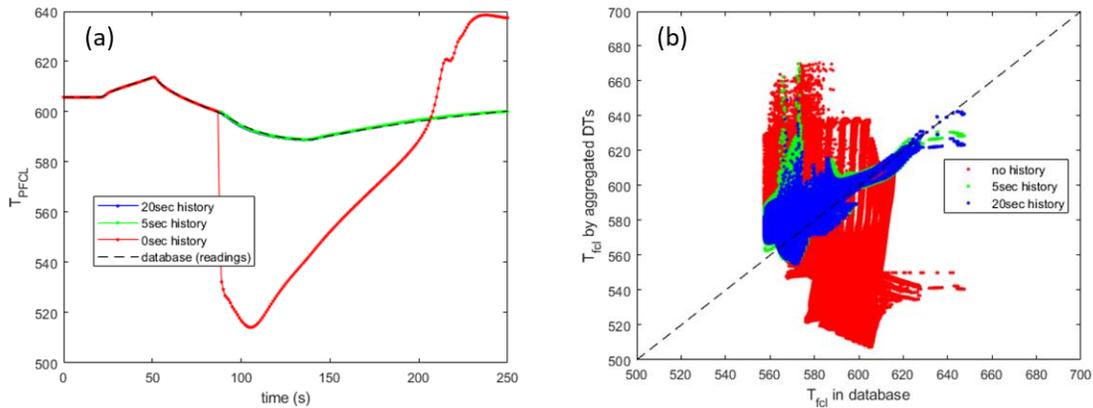

Figure 8: Comparisons of DT-P predictions for PFCL temperature with 0 s, 5 s, and 20 s history for (a) single transient and (b) all 5,000 transients.

In addition to the history information, the previous study [13] showed that the model fit hyperparameters and the coverage between training and target data have significant impacts on the performance of the DT-P. Table 8 presents a list of sources of uncertainty and their relative impacts on DT-P's accuracy. Each model fit parameter is sampled independently from the designated parameter range, and the strength of the correlation is quantified by the PCC between the DT-P errors and the sampled parameters. There is a strong correlation when the PCC absolute value is greater than 0.5. We found that RNN-based DT-P errors are strongly correlated with the sequence length, batch size, and regularization weights in the designated ranges. The data coverage between the training and target data is quantified by symmetric K-L divergence (Eq. 8) and Hellinger distance (Eq. 10). These metrics characterize the similarity between the distribution of training data and the distribution of target data, where a smaller divergence or distance indicates a similar target to the training distribution. DT-P errors grow when the target data distribution becomes different from the training data distribution.

Table 8: A list of sources of uncertainty, parameter ranges, and findings for RNN-based DT-P.

| Category | Sources of uncertainty | Parameter range | Selected value | Findings |
|---|---|---|---|---|
| **Model fit** | | | | |
| | Sequence length | [5–20] | 14 | DT-P errors are strongly affected by sequence length in the designated range, where longer lengths indicate smaller errors. However, a long sequence could degrade the training efficiency. |
| | Batch size | [100–600] | 512 | DT-P errors are strongly affected by batch size in the designated range, where larger batch sizes result in smaller errors. However, a large batch size could degrade the generalization capability. |
| | Number of neurons | [30–50] | 30 | DT-P errors are slightly affected by the number of neurons in the designated range. |
| | Regularization weight | [0–$10^4$] | 0 | DT-P errors are strongly affected by regularization weight in the designated range, where smaller weights indicate smaller errors. However, small regularizations could greatly limit the generalization capability. |
| **Scope compliance** | | | | |
| | Data coverage | Symmetric K-L divergence; Hellinger | | DT-P errors are strongly affected by data coverage by mutual information metrics. Larger mutual information between the training and target domain suggests smaller DT-P errors. |

In addition to sensitivity analysis, we applied a hyperparameter optimization method, named sequential model-based optimization (SMBO) [28]. Meanwhile, the optimization results are compared against the selected value from sensitivity analysis. In this study, a tree-structured Parzen estimator is used to build an empirical correlation between the target function, defined as the DT-P prediction errors for $T_{PFCL}$, and the hyperparameter space, defined by parameter importance analysis from the sensitivity study. 100 samples are drawn from the designated parameter range as shown in Table 8. Table 9 compares the SMBO-selected hyperparameters against the selected values from sensitivity analysis. Highly correlated hyperparameters, including sequence length, batch size, and regularization weights, have similar results, while hyperparameters, including the number of layers, number of hidden neurons, learning rates, and validation patience have different values. We obtained the RMSEs in predicting the PFCL temperature $T_{PFCL}$ and the core power rates $\dot{q}_{core}$ for all three databases. For DB2, from which the training data is drawn, the DT-P with SMBO selected hyperparameters has smaller errors than sensitivity analysis. For DB1 and DB3, the prediction errors from sensitivity analysis are smaller. Since DB1 and DB3 have different distributions than the training data, the model with hyperparameters selected by sensitivity analysis tends to have better scalability, while the SMBO-based DT-P is overfitting. As a result, hyperparameter optimization techniques depend heavily on the selection of objective function, and an objective function that does not enforce the machine learning generalization capability could aggravate the overfitting issue.

Table 9: Comparison of SMBO selected hyperparameters against the selected values from sensitivity analysis.

| Hyperparameter | SMBO selected | | | Selected value from sensitivity analysis | | |
|---|---|---|---|---|---|---|
| Sequence length | 14 | | | 14 | | |
| # of hidden neurons | 40 | | | 30 | | |
| # of layers | 4 | | | 2 | | |
| Learning rates | 0.0055 | | | 0.001 | | |
| Batch size | 550 | | | 512 | | |
| Validation patience | 80 | | | 40 | | |
| Regularization weight | $5\times 10^{-6}$ | | | 0 | | |
| **Prediction errors** | DB1 | DB2 | DB3 | DB1 | DB2 | DB3 |
| ($T_{PFCL}$) °C | 6.95 | 1.18 | 3.08 | 4.43 | 1.70 | 2.62 |
| ($\dot{q}_{core}$) MW | 2.85 | 0.38 | 1.35 | 1.44 | 0.69 | 0.92 |

In this section, a DT-P with acceptable uncertainty is developed based on the selected hyperparameters from the sensitivity analysis. Compared to the FNN-based implementation, the RNN model can make multistep predictions for the entire transient of selected state variables and thus can provide more information to better support decision-making. The scalability of the DT-P development and assessment process is demonstrated by using the same formulation and workflow to implement both RNN-based and FNN-based DT-P, which results in comparable prediction errors for the quantity of interest in the training domain. However, outside the training domain, the RNN-based DT-P has smaller uncertainty than FNN-based DT-P when they are used to predict data points. Through sensitivity analysis, this study identifies strong correlations between the DT-P errors and the selected hyperparameters, while the final selection of hyperparameters is confirmed by an SMBO hyperparameter optimization technique. We continue to observe a strong correlation between the DT-P errors and the data coverage conditions. All observations indicate consistent and scalable methods and techniques for DT-P.

### 5.3. Strategy Inventory

The strategy inventory aims to find all available mitigation strategies based on the current diagnosed state of the system. The workflow of strategy inventory is discussed in [8], and the same scheme is used in this study. Since the RNN-based prognosis requires the transients of pump torques, the strategy inventory needs to predict the torque curves for both primary pumps and to assimilate the prediction with measurement histories. In this study, linear curves are generated for each combination of malfunctions and mitigations according to the end magnitudes of malfunctioned and controllable primary pumps. Eq. 14 and Eq. 15 show the piecewise functions for torque curves of PSP#1 and PSP#2.

$$\tau_1(t) = \begin{cases} \tau_m(t), & t < t_{rcmd} \\ \tau_m(t_{rcmd}) - \dfrac{\tau_m(t_{rcmd}) - [\tau_1]_{end}}{[t_1]_{end} - t_{rcmd}}(t - t_{acc}), & t_{rcmd} \leq t < [t_1]_{end} \\ [\tau_1]_{end}, & t_{end} \leq t < T \end{cases} \quad \text{Eq. 14}$$

$$\tau_2(t) = \begin{cases} \tau_0, & t < t_{trip} \\ \tau_0 + \dfrac{[\tau_2]_{end} - \tau_0}{[t_2]_{end} - t_{trip}}(t - t_{trip}), & t_{trip} \leq t < [t_2]_{end} \\ \tau_{end}, & [t_2]_{end} \leq t < T \end{cases} \quad \text{Eq. 15}$$

$t_{rcmd}$ is the recommendation time when NAMAC is activated and the strategy inventory is called to predict the torque transients for both pumps. $\tau_1(t)$ represents the torque curve of PSP#1, which is assumed to have malfunctioned. $\tau_2(t)$ represents the curves of PSP#2, which is controlled to mitigate the partial LOF induced by PSP#1 malfunction. Before $t_{rcmd}$, the PSP#1 torque curve is built based on sensor measurements $\tau_m(t)$. After $t_{rcmd}$, the torque curve is predicted based on the last sensor measurement at $t_{rcmd}$, the malfunction starting time $t_{acc}$, ending time $[t_1]_{end}$, and final operating condition represented by the final torque $[\tau_1]_{end}$. For the PSP#1 malfunction, these variables are supposed to be determined by DT-D. For the demonstration in this study, we assumed the values of these variables. To mitigate the situation, PSP#2 is ramped up from the nominal torque $\tau_0$ by increasing the torque to a final torque $[\tau_2]_{end}$, which starts from $t_{trip}$ and ends at $[t_2]_{end}$. $T$ represents the end of the prediction regime, and the entire transient lasts for 250 s.

Figure 9 shows the torque trajectories for PSP#1 and PSP#2, respectively. In this example, the strategy inventory is activated at $t_{rcmd}$ = 37.8sec, when the PSP#1 is in the middle of ramping down due to malfunctions. The PSP#1 starts the malfunction at 20 s, and we assumed by the diagnosis that it will reach and stay at $[\tau_1]_{end}$. = 494.23 $N \cdot m$ after $[t_1]_{end}$=70sec. As a result, based on the sensor measurements before $t_{rcmd}$, Eq. 14 describes a linear piecewise curve from $t_{rcmd}$ to the end of the transient. However, because of the nonlinear interactions between pump torque and speed, the PSP#1 torque is actually ramping down faster than expected, which results in a mismatch in gradients (ramping-down speeds) between the sensor measurements and strategy inventory prediction. As a result, we observed a discrepancy between the real PSP#1 torque curve from the database and the strategy inventory predictions. In response to the malfunction, one of the mitigation strategies proposed by the strategy inventory is to ramp up PSP#2 immediately from $t_{rcmd}$ until it reaches the final torque $[\tau_2]_{end}$ = 750.30 $N \cdot m$ at 87.8 s. Based on this information, Eq. 15 results in a linear piecewise function from $t_{rcmd}$ to the end of the transient. Again, because of the nonlinearity in pump behaviors, a small discrepancy is observed at the beginning of the ramping up and the final operating torques.

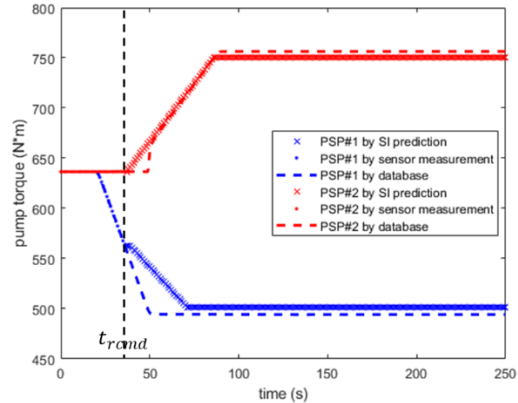

Figure 9: Transient of PSP torques for PSP#1 in blue and PSP#2 in red. The cross marks represent sensor measurements before $t_{rcmd}$, while the dots are predictions by the strategy inventory. The dash lines are real torque curves from the database. SI stands for strategy inventory.

Figure 10 compares the strategy inventory predictions for (a) PSP#1 and (b) PSP#2 torques against the real values from the databases, respectively. Most predictions fall onto the 45° line in Figure 10 (b), which indicates that Eq. 15 has good accuracy in predicting the torque of PSP#2. In Figure 10 (a), predictions fall outside the 45° line when the torques are small, and thus predictions by Eq. 14 have a lower accuracy as the torque gets close to zero. The RMSE for both predictions (5.89 N·m and 6.53 N·m) are found below the 5% uncertainty bounds.

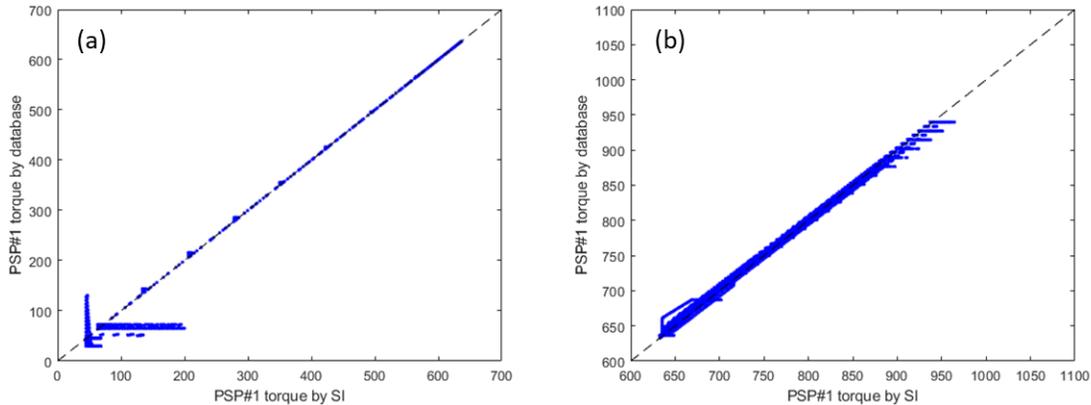

Figure 10: Comparisons of strategy inventory predictions for PSP#1 (a) and PSP#2 (b) torques against the real values from the database.

Overall, a strategy inventory model is developed with an acceptable uncertainty in predicting the trajectories of both PSPs. The strategy inventory is tested on both DB2 and DB3, while the errors in predicting PSP#1 torques are 5.89 N·m and 11.56 N·m, respectively. Both are bounded by 5% uncertainty bounds (<32 N·m). Compared to the demonstration of the baseline strategy inventory, the architecture, workflow, development, and assessment process are scalable from the NAMAC proof of concept to the complex LOF scenarios. However, the model errors increase in the low-torque regions, where pump behaviors are highly nonlinear because of the potential for the unbalanced flow to cause reverse flow and possibly a locked rotor condition.

### 5.4. Utility Function for Strategy Assessment

To determine whether a mitigation strategy and its effects are preferred, a reward function is designed to evaluate mitigation strategies based on the predicted action effects in the target domain [29]. The workflow of strategy assessment is discussed in [8], and the same scheme is used in this study. For multi-attribute decision problems, the reward region is used with the upper and lower boundaries for each operational goal [30]. In this study, three reward regions are assigned to three state variables based on their operational goals, respectively. Figure 11 shows the assignment of "Best," "Good," and "Bad" reward regions based on the ranges of (a) PFCL temperature $T_{PFCL}$, (b) fractional variation of core power rates, and (c) fractional variation of pump torques. The first decision attribute is based on the PFCL temperature, and the operational goal is to maintain the fuel temperature as close to the nominal temperature as possible while staying away from the operational limit (685℃) during accident scenarios. As a result, the best region has the nominal fuel temperature of 600℃ as the lower limit and 615℃ as the upper limit. The following good region has an upper limit of 685℃, while any $T_{PFCL}$ higher than 685℃ is classified as the bad region. The second attribute is the variations of core power rates, and the operational goal is to maintain core power generation at full power (nominal) during accident scenarios. The best region requires the fractional variations to be bounded by 10% of full power rates, while the good region requires the fractional variations to be bounded by 20% of full power rates. Any power variations exceeding 20% are classified into bad reward regions. The third attribute is the variation of pump torques, and the operational goal is to maintain the torques of PSP at their best efficiency point (BEP) and avoid significant overspeed of PSP during accident scenarios.

To reduce pump wear and increase pump reliability, the pump should be operated around the BEP, known as preferred operating region (POR), such that flow enters and leaves the pump with a minimum amount of flow separation, turbulence, and other losses. Meanwhile, a wider range of operations outside the POR is also acceptable. In this range, known as the allowable operating region (AOR), the reliability of the pump will be affected by the internal hydraulic loads and flow-induced vibration. These impacts result in higher energy consumptions and more frequent maintenance or even replacement than expected. Moreover, there are multiple negative consequences when the pump is operated significantly away from the BEP or outside the AOR. The pump wear can be largely accelerated and result in premature failures. For example, operating pump at excessively high speed may not meet the pump's required suction head, and the pump could cavitate. However, if overspeeding is needed to mitigate an accident, e.g., maintain the structural integrity, avoid severe accidents, etc., it is a worthwhile decision since the scripted accident procedures are written primarily with safety objectives in mind. As a result, this study takes the PSP nominal torques as the BEP and requires the fractional variation of PSP torques around the nominal conditions to be less than 25%, while the good region requires the fractional variations to be less than 50%. Any torque variations exceeding 50% are classified into bad reward regions.

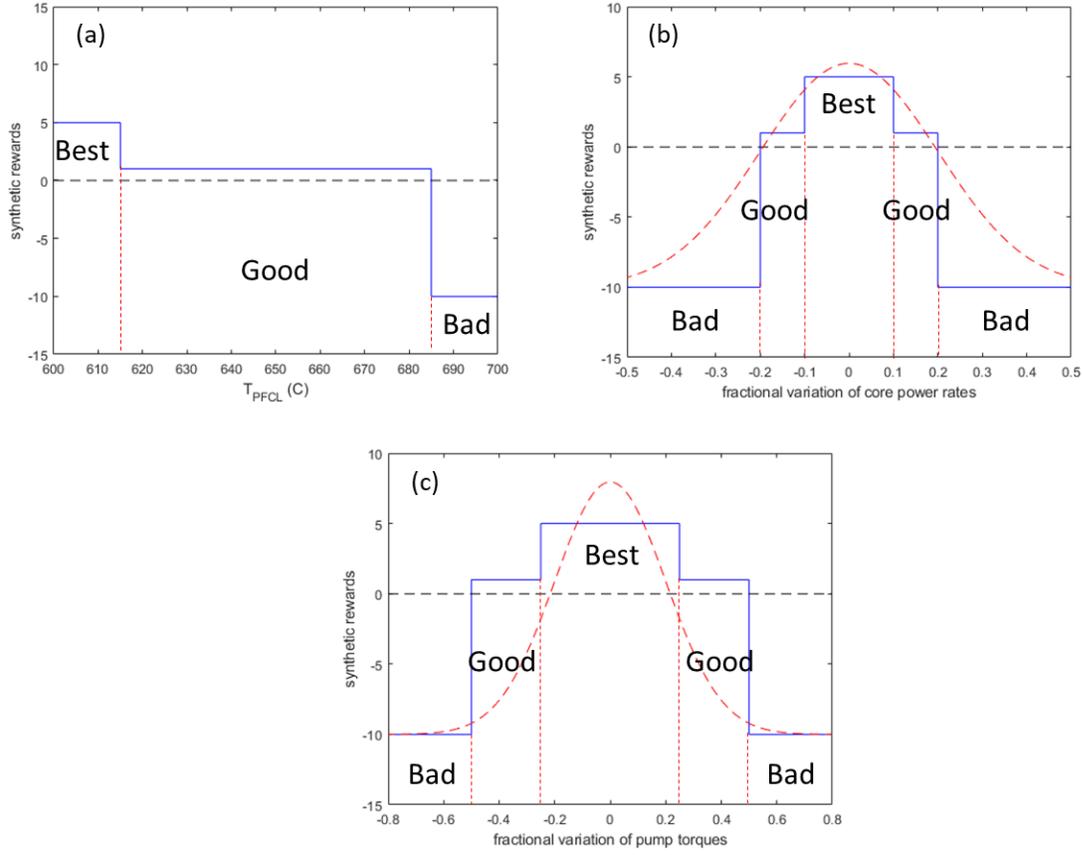

Figure 11: Assignment of reward regions based on the distribution of (a) PFCL temperature $T_{PFCL}$, (b) fractional variation of core power rates, (c) fractional variation of pump torques.

To calculate the total rewards based on the operational goal and reward regions, three synthetic rewards are assigned to each region (Eq. 16–Eq. 18). We further assumed that the rewards are accumulating along with the temporal transient and that the rewards from different attributes are additive. As a result, Eq. 19 is used to determine the total rewards $R_T$ with the weighted sum method, while coefficients $a$, $b$, and $c$ suggest the relative weights from each attribute. In this study, a uniform weight is assigned, while in real operation, the weight should depend on the relative importance of each attribute, which can be mathematically related to decision-makers' preference function [31].

$$R_{best} = 5 \quad \text{when} \quad \begin{aligned} &615°\text{C} \geq T_{PFCL} > 600°\text{C} \\ &10\% \geq |\Delta \bar{\dot{q}}_{core}| \\ &25\% \geq |\Delta \bar{\tau}_2| \end{aligned} \quad \text{Eq. 16}$$

$$R_{good} = 1 \quad \text{when} \quad \begin{aligned} &685°C \geq T_{PFCL} > 615°C \\ &20\% \geq |\Delta \overline{\dot{q}_{core}}| > 10\% \\ &50\% \geq |\Delta \overline{\tau}_2| \end{aligned} \quad \text{Eq. 17}$$

$$R_{bad} = -10 \quad \text{when} \quad \begin{aligned} &T_{PFCL} > 685°C \\ &|\Delta \overline{\dot{q}_{core}}| > 20\% \\ &|\Delta \overline{\tau}_2| > 50\% \end{aligned} \quad \text{Eq. 18}$$

$$R_T = a \int_0^{250} R_{T_{PFCL}} \cdot dt + b \int_0^{250} R_{\dot{q}_{core}} \cdot dt + c \int_0^{250} R_{\tau_2} \cdot dt \quad \text{Eq. 19}$$

$$a + b + c = 1 \quad \text{Eq. 20}$$

The weighted sum method is the most common approach to multi-objective and multi-attribute decision-making problems. Since weights have significant impacts on the final rewards, many systematic methods have been developed to select weights, and surveys of these methods can be found in [32]. However, previous research [33] reported that the weighted sum approach is sometimes difficult to obtain points on non-convex portions of the Pareto optimal set in the criterion space. There is also an issue regarding *a priori* selections of weights because the weights must be functions of the original objectives, not constants, such that a weighted sum can mimic a preference function accurately [34]. However, the goal of this study is not to optimize the design and use of decision-making methods but to demonstrate the capability of NAMAC in making transparent and flexible recommendations. A comprehensive survey of different multi-objective optimization methods for engineering problems can be found in [35]. Figure 12 shows the accumulation of rewards along with the time transient from the three attributes represented by different colors. The rewards from $T_{PFCL}$ are the major contributors to total rewards since the temperature is maintained below 615°C, while both power variations and torque variations go into the "good" region with smaller rewards accumulation rates.

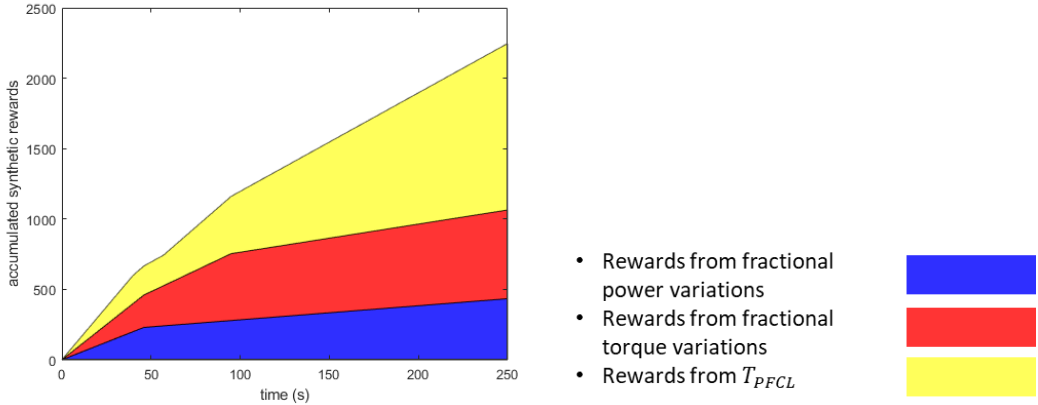

Figure 12: The accumulation of rewards from different attributes.

Overall, the strategy assessment component in this study is developed with a scalable architecture and function from the NAMAC proof of concept to the complex LOF scenarios. The multi-attribute decision-making scheme is able to account for reactor safety, power generation, and component reliability at the same time. Although the rewards function is constructed with synthetic rewards and arbitrarily assigned operational regions, the relative contributions of each attribute and region are interpretable and can be easily visualized. Moreover, the relative weights of each objective can be modified such that global criteria and preference on reactor safety, performance, component reliability, and maintenance can be easily incorporated.

## 5.5. Discrepancy Checker

To avoid severe consequences due to unreliable NAMAC recommendations, the discrepancy checker is needed to evaluate the accuracy of DTs and to decide whether safety-minded control actions (e.g., reactor trip) are needed. In this study, the discrepancy between the expected transients in response to the NAMAC recommendation and the real reactor conditions is evaluated. Figure 13 shows the workflow of the discrepancy checker.

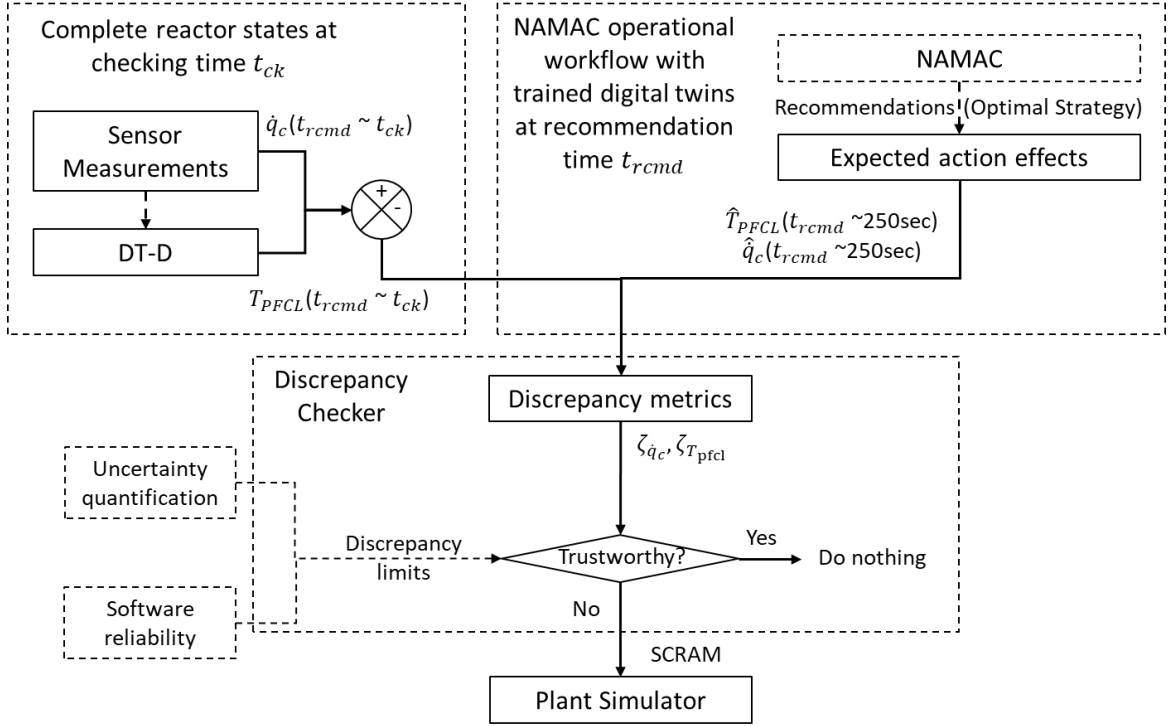

Figure 13: Workflow of discrepancy checker.

The expected action effects are predicted by DT-P at the recommendation time $t_{rcmd}$, including the transients of fuel centerline temperature $\hat{T}_{PFCL}$ and core power rate $\hat{\dot{q}}_c$. Next, at the designated checking time $t_{ck} = \{t_{ck_1}, \dots, t_{ck_i}\}$, the core power rate $\dot{q}_c$ measured by sensors and the PFCL temperature $T_{PFCL}$ inferred by the DT-D are compared against the corresponding expected action effects. Eq. 21 and Eq. 22 show the calculation of discrepancy factors $\zeta_{\dot{q}_c}$ and $\zeta_{T_{\text{pfcl}}}$, where the errors $\varepsilon_{\dot{q}_c}$ and $\varepsilon_{T_{\text{pfcl}}}$ are calculated by Eq. 11. These factors are compared against the discrepancy limits. Reference [19] discussed the importance of uncertainty quantification and software reliability analysis to decide the reliability and trustworthiness of NAMAC recommendations. Considering the complexity of trustworthiness and reliability assessment and the heterogeneity of supporting materials, a formalized argumentation framework is recommended in [36] to argue the assurance of NAMAC for the intended uses. Since the goal of this study is not to propose a trustworthiness assessment framework, a 10% discrepancy limit is assigned for demonstration purposes. If either of these factors is lower than 10%, no action is recommended by discrepancy checker, and the reactor will continue operations. Otherwise, a SCRAM command is sent to the plant simulator to shut down the reactor.

$$\zeta_{\dot{q}_c} = \frac{\varepsilon_{\dot{q}_c}}{\dot{q}_c(0)} \qquad \text{Eq. 21}$$

$$\zeta_{T_{\text{pfcl}}} = \frac{\varepsilon_{T_{\text{pfcl}}}}{T_{\text{pfcl}}(0)} \qquad \text{Eq. 22}$$

Overall, the discrepancy checker component is developed with a scalable architecture and function from the NAMAC proof of concept to the complex LOF scenarios. The discrepancy checker decides the reliability of NAMAC recommendations based on the uncertainty of NAMAC predictions. We propose a simplified discrepancy checking scheme in this study, while the workflow should be scalable to more complex scenarios in a transparent and consistent manner.

## 6. DEMONSTRATION OF NAMAC

To demonstrate the use of NAMAC, all DTs, strategy inventory, strategy assessment, and the discrepancy checker are connected by the operational workflow (Figure 2). In addition, we built a plant simulator to simulate the thermal-hydraulic behaviors for the primary side of EBR-II. To avoid inconsistency and reduce uncertainty, the same GOTHIC simulator (GSIM) for knowledge base generation is used as the EBR-II plant simulator. An I&C interface is built into GOTHIC to inject mitigation strategies (i.e., PSP torque control and reactor shut down) while the simulation is running. A designated malfunction of PSP is injected as the safety challenges. Later, when NAMAC is activated for making a recommendation at recommendation time $t_{rcmd}$, the plant simulator is paused until the NAMAC system derives a recommendation. First, unobservable SSFs, including the PFCL and peak cladding temperature, are continuously predicted by the DT-D based on the sensor information. Figure 14 compares the DT-D predictions against the simulated values by GSIM, where the malfunction is injected at 10 s and the GSIM is paused at 20 s for recommendations.

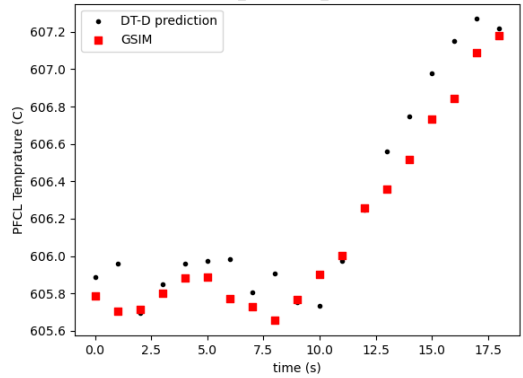

Figure 14: Comparisons of DT-D predictions against the GSIM predictions.

Next, based on the collected information and constraints in mitigation strategies, the strategy inventory finds all available mitigation strategies and predicts the torque transients for both PSPs. The strategy in found by comparing the predicted fuel temperature by DT-D against a list of reference temperatures, each of which corresponds to a final torque speed $[\tau_2]_{end}$ and a start time for ramping up $t_{trip}$. Figure 15 shows 245 available torque transients from 300 options.

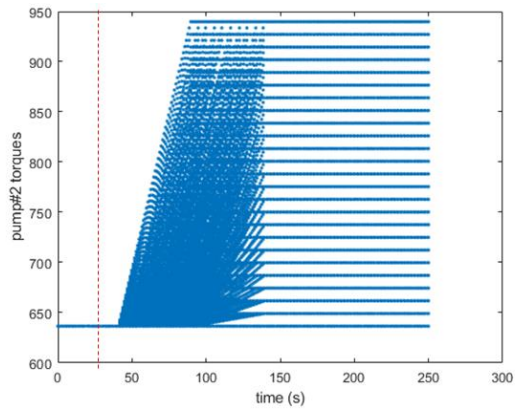

Figure 15: Predicted torque transients for PSP#2 by the strategy inventory. The dashed line is the recommendation time. Data before the dashed line is obtained from sensors, data after is predicted by Eq. 15.

Next, the DT-P predicts the future behaviors of the reactor for each optional mitigation strategy based on the predicted transients and the history of reactor states (Figure 16).

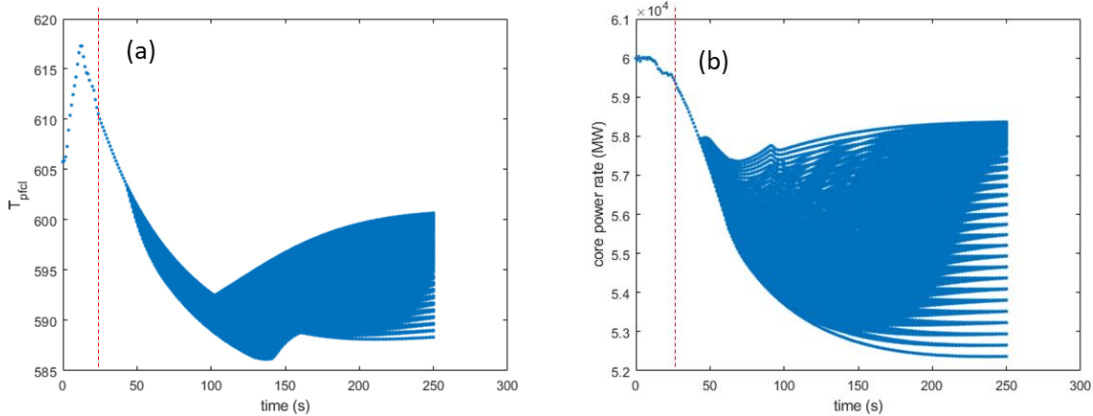

Figure 16: Predicted transients for (a) fuel centerline temperature and (b) core power rates. Dash line is the recommendation time $t_{rcmd}$. Data before $t_{rcmd}$ are (a) obtained from DT-D (b) obtained from sensors. Data after $t_{rcmd}$ are predicted by DT-P.

Next, based on the reward region and operational goal, the strategy assessment calculates the total rewards for each predicted transient and recommends the optimal action that results in the highest rewards. Figure 17 plots the contour of rewards based on Eq. 16–Eq. 20 with uniform weights to each reward attribute. The highest rewards are 1,903 where the PSP#2 is ramped up to 716.15 $N \cdot m$ starting from 68 s. The predicted maximally reachable fuel temperature is 608.7°C, while the predicted maximum and minimum power variations are 99.0021% and 101.9% to the nominal power, respectively.

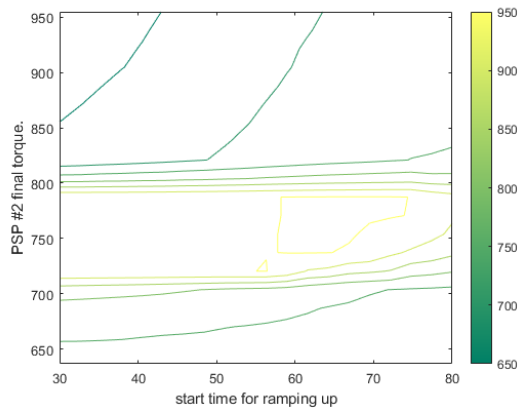

Figure 17: Contour plot of total rewards concerning the final torque speed $[\tau_2]_{end}$ and the start time for ramping up $t_{trip}$.

If the operator accepts the NAMAC recommendation, the mitigation strategy with the highest rewards is injected into the I&C system of the plant simulator, and the plant simulator continues simulating the event. Next, the discrepancy checker is activated at checking time $t_{ck}$ = 100 s and 200 s separately to determine if the reactor responses are as we expected. Characterized by two discrepancy factors (Eq. 21 & Eq. 22), if either of them exceeds the limits (10%), a reactor trip command is sent to the I&C interface of the plant simulator. Figure 18 compares the DT-P predictions against sensor measurements from GSIM for core power rates. The RMSEs are 0.36 MW (at $t_{ck}$ = 100 sec) and 0.54 MW ($t_{ck}$ = 200 sec), and the discrepancy factors by Eq. 21 are 0.60% and 0.88%, respectively. Meanwhile, another discrepancy factor for the PFCL temperature by Eq. 22 is 0.48% and 0.43%, and both factors satisfy the criteria. As a result, the discrepancy checker suggests continuing the reactor operations with NAMAC recommendations.

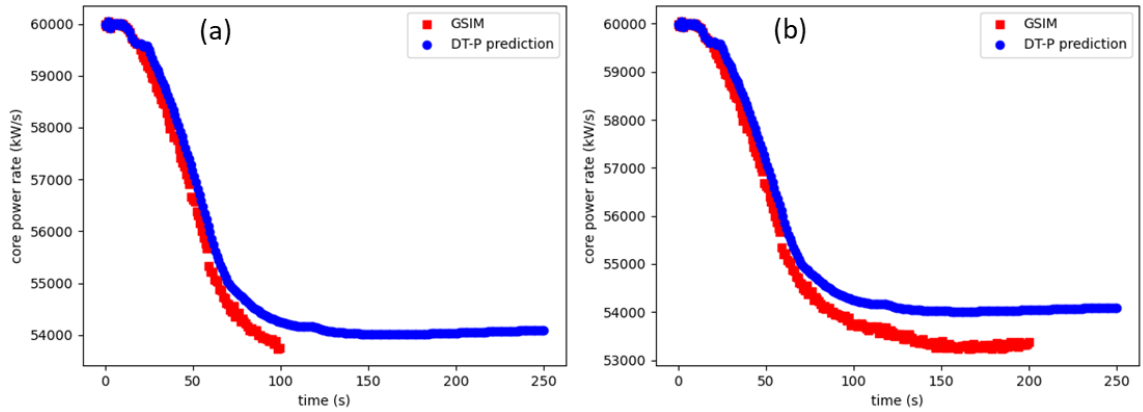

Figure 18: Comparisons of DT-P predictions against sensor measurements from GSIM for core power rates at (a) $t_{ck}$ = 100 s and (b) $t_{ck}$ = 200 s.

In addition to testing within the training domain, we also tested the NAMAC system with 46 different instances of PSP#1 malfunction. The malfunction is specified by the speed of torque reductions in percentage/s as malfunction speed and the percentage of torque losses over the nominal torque in percentage as malfunction magnitude. Meanwhile, the NAMAC prediction errors are measured by the RMSE in predicting the recommended transient of PFCL temperature and core power rates. Figure 19 plots the surface of RMSEs against the malfunction speed and the malfunction magnitude. The errors grow significantly when there are 100% losses in torques and when the torque reduces faster than 5%/s. As discussed in Section 3, this is mainly caused by the nonlinear behaviors of the plant simulator when the torque differences between two PSPs are greater than 500 N·m. However, when comparing it to the baseline NAMAC with FNN-based DTs [13], the NAMAC in this study has smaller errors in much wider malfunction instances. According to the limit in discrepancy checker, all $\zeta_{T_{\text{pfcl}}}$ by Eq. 21 is smaller than 10%, while $\zeta_{T_{\text{pfcl}}}$ in three cases with 100% malfunction magnitude and 10%/s malfunction speed exceeds 10% limit.

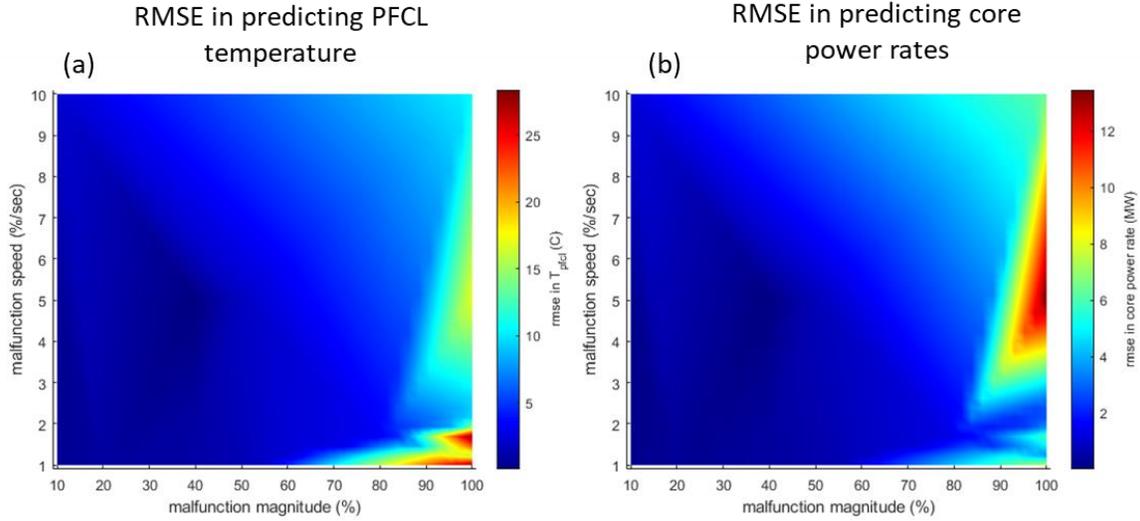

Figure 19: Surface plot of RMSE for (a) PFCL temperature $T_{FPCL}$ and (b) core power rates against real values from GSIM. The surface is constructed against the malfunction speed in %/s and the malfunction magnitude in %.

To demonstrate the impacts of NAMAC prediction errors on the decision-making, the NAMAC decision error $\varepsilon_{R_i}$ is calculated by Eq. 23.

$$\varepsilon_{R_i} = \frac{\sqrt{\frac{1}{N}\sum_{j=1}^{N}(\hat{R}_j - R_j)^2}}{R_i} \qquad \text{Eq. 23}$$

where $R_i = \{R_{T_{PFCL}}, R_{\dot{q}_{core}}, R_{\tau_2}]$ corresponds to rewards by the range of PFCL temperature, the fractional variation of pump torques, and the fractional variation of core power rates. $\hat{R}_j$ are rewards determined based on NAMAC's prediction for the recommended mitigation strategies at time step $j$, while $R_j$ are rewards calculated by real responses from the plant simulator at the same time step. For all three attributes, rewards errors from pump torques are smaller than 20% and negligible. Figure 20 plots the surface of scaled NAMAC decision errors $\varepsilon_{R_i}$ for (a) PFCL temperature ($R_{T_{PFCL}}$), and (b) the fractional variation of pump torques ($R_{\dot{q}_{core}}$). All surfaces are constructed against the malfunction speed and the malfunction magnitude. For $\varepsilon_{R_{\text{Tpfcl}}}$, the errors grow when the malfunction magnitude gets close to 100%, which is caused by the nonlinear behaviors with large differences in PSP torques. If a 20% limit is assigned to the rewards errors, 75% of cases satisfy the requirements for $R_{T_{PFCL}}$. The differences in $R_{\dot{q}_{core}}$ increase when the malfunction magnitude is higher than 50%, and only half of the 46 cases fall below the 20% limit. This is mainly caused by the small power variation (10%, 20%) limits in Eq. 16–Eq. 18. When the limit is relaxed to 15% for the best region and 30% for the good region, the average $\varepsilon_{R_{\dot{q}}}$ is reduced from 1.07 to 0.004. The focus of this study is not to find the optimal reward region and value but to demonstrate the capability of NAMAC in making a multi-attribute decision. However, a further sensitivity analysis needs to be conducted for the assignment of rewards and relative weights, together with the form of rewards function. Moreover, to prevent severe consequences due to NAMAC discrepancy, the discrepancy checker will be automatically activated and send a scram signal if large deviations are observed, especially for the severe LOFs (100% loss) with rapid torque reductions in 10 sec.

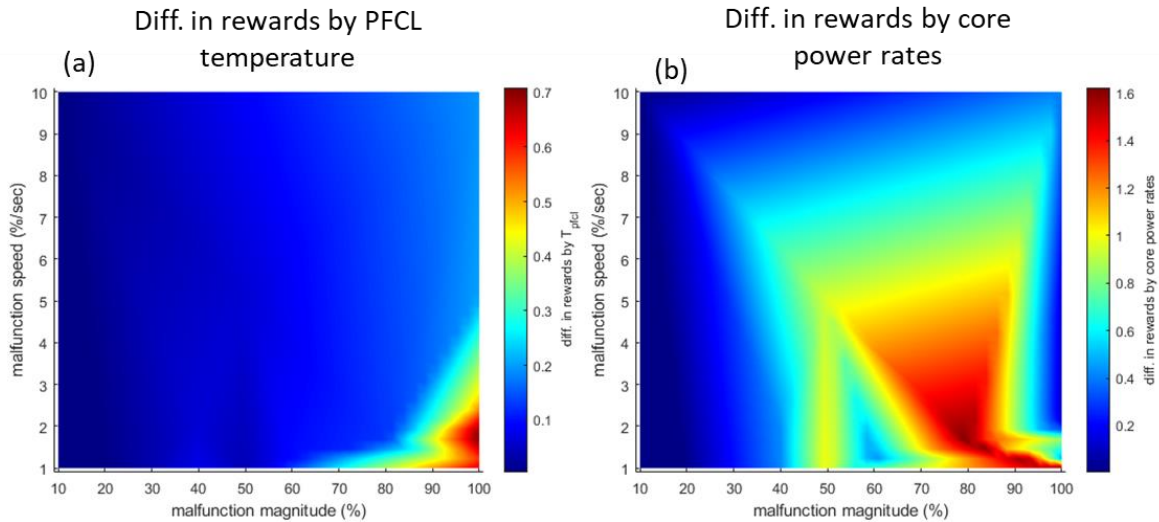

Figure 20: Surface plot of rescaled NAMAC decision error $\varepsilon_{R_i}$ for (a) PFCL temperature and (b) the fractional variation of pump torques. The surfaces are constructed against the malfunction speed in %/sec and the malfunction magnitude in %.

Overall, this study demonstrates the capability of NAMAC in making recommendations with complex decision-making during complex LOF scenarios. The uncertainty of NAMAC is evaluated in a class of LOF scenarios with various magnitudes and speed of malfunction, while most cases are found bounded by the uncertainty requirements. Meanwhile, the scalability of NAMAC is tested with a consistent operational workflow, reactor I&C interface, and recommendations from simplified (proof of concept) to complex scenarios.

## 7. CONCLUSION

To adapt NAMAC to a broader range of accident scenarios, this study develops a NAMAC for recommending mitigation strategies with the highest rewards during various degrees of LOF scenarios. The implemented coupled NAMAC-simulator system provides a platform for testing different DT modeling techniques, the construction of multi-attribute and user-defined reward functions, and the demonstration of autonomous control systems in different issue spaces.

The improved design of NAMAC and DT shows robustness to broader issue spaces, the capability of multi-attribute decision-making, and flexibility to complex scenarios and systems. Three databases are generated, and a list of state variables is selected from the I&C and DT requirements. Based on the extracted training, validation, and testing database, the DT-Ds and DT-Ps are implemented with RNN. For DT-D implementation, the RNN-based diagnosis is more robust against issue-space coverage and sensor noises. A sensitivity analysis also shows that the DT-D errors strongly depend on the batch size, regularization weight, and data coverage quantified by symmetric K-L divergence and Hellinger distance. For DT-P implementation, an RNN-based prognosis is developed for multistep predictions, which provides richer information than the FNN models. The sensitivity analysis shows that the DT-P errors strongly depend on the sequence length, batch size, regularization weight, and data coverage. The selected hyperparameters from the sensitivity analysis are compared against an SMBO result, where similar values are found for highly dependent hyperparameters. However, the selected values from sensitivity analysis tend to have smaller DT-P errors in predicting databases with different distributions than the training database. A model-based strategy inventory is developed to predict the trajectory of PSP torques based on the malfunction and mitigation magnitudes. However, the gaps between linear models and nonlinear pump behaviors could result in large prediction errors. To rank mitigation strategy options based on their action effects, a multi-attribute reward function is developed with synthetic rewards and three operational regions. At last, we implemented a discrepancy checker with two discrepancy factors, which sends reactor trip signals when any factors exceed the limits.

To demonstrate the NAMAC, all implemented components are connected by an operational workflow and coupled with a GOTHIC EBR-II simulator. In addition to the basic demonstration within the training domain, 46 tests are performed with various speeds and magnitudes of malfunctions. The NAMAC errors for predicting the recommendation action effects (on average 96% among all cases) and the recommendation rewards (on average 67% among all cases) meet the requirements. However, the errors increase when the malfunction magnitudes are close to 100%. Specifically, the errors of rewards by core power rates increase when the malfunction magnitude is higher than 50%, which is mainly caused by the strict power variation (10%, 20%) requirements in the reward functions. A relaxation in power variation limits could greatly reduce the errors of rewards.

In addition to NAMAC's uncertainty in a broad issue space, this study demonstrates the scalability of NAMAC's knowledge base, components, and system from the proof of concept to the complex LOF scenarios. In the future, on one hand, we will continue enhancing the diversity and complexity of NAMAC applications. More specifically, we will apply NAMAC to broader issue spaces with different initiating events and combinations of events. Meanwhile, we plan to test NAMAC on different advanced reactor designs and use cases, including power controls, normal operations, etc. On the other hand, we stress the importance of NAMAC robustness and trustworthiness, including the uncertainty and scalability of separate components and integral system. The goal is to enhance the confidence and transparency for NAMAC's operation and regulatory acceptance.

## ACKNOWLEDGMENT

This work is performed with the support of the ARPA-E MEITNER program under the project entitled "Development of a Nearly Autonomous Management and Control System for Advanced Reactors" and using a GOTHIC license provided by Zachry Nuclear Engineering, Inc.

GOTHIC incorporates technology developed for the electric power industry under the sponsorship of the Electric Power Research Institute.

The authors would also like to acknowledge the comments and suggestions from Dr. Cristian Rabiti and Dr. Paolo Balestra of Idaho National Laboratory; Dr. Min Chi of North Carolina State University; Mr. John Link of Zachry Nuclear Engineering; Dr. Botros Hanna and Dr. Son Tran of New Mexico State University; Dr. David Pointer, Dr. Sacit Cetiner, and Dr. Birdy Phathanapirom of Oak Ridge National Laboratory; and Dr. Carol Smidts of Ohio State University. Work at Idaho National Laboratory was performed under Department of Energy Idaho Operations Office Contract DE-AC07-05ID14517.